\documentclass[runningheads]{llncs}
\usepackage{graphicx}
\usepackage[table,xcdraw,dvipsnames]{xcolor}
\usepackage{tikz}
\usepackage{comment} 
\usepackage{amsmath,amssymb} % define this before the line numbering.
\usepackage{color}
\usepackage{pdfpages}
\usepackage{multido}

\newcommand{\psm}{PSMNet \cite{Chang_2018_CVPR}}
\newcommand{\iresnet}{iResNet \cite{Liang_2018_CVPR}}

\newcommand{\stereodepth}{Stereodepth}

\newcommand{\adcensus}{Block Matching}
\newcommand{\sota}{state-of-the-art}

\newcommand{\gt}{ground-truth}
\newcommand{\eg}{e.g.}
\newcommand{\ie}{i.e.}
\newcommand{\etal}{et al.}
\newcommand{\tradeoff}{trade-off}
\newcommand{\ee}{end-to-end}
\newcommand{\illpose}{\textit{ill-posed}}
\definecolor{LightCyan}{rgb}{0.88,1,1}
\definecolor{LightYellow}{rgb}{1,1,0.7}
\definecolor{LightGreen}{rgb}{0.4, 1, 0.6}

\definecolor{lower}{RGB}{232, 161, 148}
\definecolor{upper}{RGB}{148, 187, 232}

\newcommand{\stereo}{\mathcal{S}}

\newcommand{\filter}{\mathcal{F}}
\newcommand{\dispL}{\mathcal{D}}
\newcommand{\imageL}{\mathcal{I}^L}
\newcommand{\imageR}{\mathcal{I}^R}
\newcommand{\dispD}{\mathcal{D}^\mathbb{O}}
\newcommand{\dispP}{\mathcal{D}^\mathbb{P}}

\begin{document}
% \linenumbers
\pagestyle{headings}
\mainmatter
\def\ECCVSubNumber{1420}  % Insert your submission number here

\title{Reversing the cycle: self-supervised deep stereo through enhanced monocular distillation}

\titlerunning{Reversing the cycle} 
\authorrunning{F. Aleotti \etal{}} 

\author{
Filippo Aleotti\thanks{Joint first authorship}\inst{1}%\orcidID{0000-0002-8911-3241}
\and
Fabio Tosi$^\star$\inst{1}%\orcidID{0000-0002-6276-5282} 
\and 
Li Zhang\thanks{Work done while at University of Bologna.}\inst{2}
\and \\
Matteo Poggi\inst{1}%\orcidID{0000-0002-3337-2236} 
\and
Stefano Mattoccia\inst{1}%\orcidID{0000-0002-3681-7704}
}
\institute{University of Bologna, Viale del Risorgimento 2, Bologna, Italy \and
China Agricultural University, Beijing, China}
%******************
\maketitle

\begin{abstract}

In many fields, self-supervised learning solutions are rapidly evolving and filling the gap with supervised approaches. This fact occurs for depth estimation based on either monocular or stereo, with the latter often providing a valid source of self-supervision for the former.
In contrast, to soften typical stereo artefacts, we propose a novel self-supervised paradigm reversing the link between the two. Purposely, in order to train deep stereo networks, we distill knowledge through a monocular completion network. This architecture exploits single-image clues and few sparse points, sourced by traditional stereo algorithms, to estimate dense yet accurate disparity maps by means of a consensus mechanism over multiple estimations. We thoroughly evaluate with popular stereo datasets the impact of different supervisory signals showing how stereo networks trained with our paradigm outperform existing self-supervised frameworks. Finally, our proposal achieves notable generalization capabilities dealing with domain shift issues. Code available at \url{https://github.com/FilippoAleotti/Reversing}.

\keywords{stereo matching, self-supervised learning, distillation}
\end{abstract}

\section{Introduction}
Among techniques to infer depth, stereo is an effective and well-established strategy to accomplish this task deploying two cameras. Stereo methods, at first, compute disparity by matching corresponding points across the two images and then recover depth through triangulation, determining the parameters of the stereo rig beforehand with calibration. Nowadays, deep learning architectures have outperformed traditional methods by a large margin in terms of accuracy on standard benchmarks. Nonetheless, \sota{} solutions require a large amount of data and \gt{} labels to learn how to perform \textit{matching}, \ie{} find in the other view corresponding pixels. The advent of self-supervised solutions based on image reprojection overcomes this limitation at the cost of weak performance in presence of occluded and texture-less regions, \ie{} where the matching does not occur.

In recent years, single-image depth estimation methods, in general up to a scale factor, gained ever-increasing attention. In this field, despite the \textit{ill-posed} nature of the problem, deep learning architectures achieved outstanding results as reported in the literature. By construction, a monocular method does not infer depth by matching points between different views of the same scene.
Therefore, compared to stereo approaches, monocular ones infer depth relying on different cues and thus potentially not affected by some inherent issues of stereo, such as occlusions. Even if supervision is sourced from stereo images \cite{Godard_CVPR_2017}, a set of practices suited for the specific monocular task allow networks to avoid undesired artifacts in correspondence of occlusions \cite{Godard_ICCV_2019}.
Starting from these observations, we argue that a single image method could potentially strengthen a stereo one, especially in occluded areas, but it would suffer the inherent scale factor issue.
Purposely, in this paper we prove that traditional stereo methods and monocular cues can be effectively deployed jointly in a \textit{monocular completion network} able to alleviate both problems, and thus beneficial to obtain accurate and robust depth predictions.

Our contributions can be summarized as follows: 
i) A new general-purpose methodology to source accurate disparity annotations in a self-supervised manner given a stereo dataset without additional data from active sensors. To the best of our knowledge, our proposal is the first leveraging at training time a novel self-supervised monocular completion network aimed specifically at ameliorate annotations in critical regions such as occluded areas. 
ii) In order to reduce as much as possible inconsistent disparity annotations, we propose a novel consensus mechanism over multiple predictions exploiting input randomness of the monocular completion network.
iii) The generated proxies are dense and accurate even if we do not rely on any active depth sensor (\eg{} LiDAR). 
iv) Our proxies allow for training heterogeneous deep stereo networks outperforming self-supervised \sota{} strategies on KITTI. Moreover, the networks trained with our method show higher generalization to unseen environments. 

\section{Related work}

In this section, we review the literature relevant to our work.

\textbf{Traditional and Deep Stereo.} Depth from stereo images has a longstanding history in computer vision and several hand-designed methods based on some of the steps outlined in \cite{scharstein2002taxonomy} have been proposed. For instance, a fast yet noisy solution can be obtained by simply matching pixels according to a robust function \cite{Secaucus_1994_ECCV} over a fixed window (\adcensus), while a better accuracy-speed \tradeoff{} is obtained by running Semi-Global Matching (SGM) \cite{SGM_PAMI}. Recently, deep learning proved unpaired performance at tackling stereo correspondence. Starting from matching cost computation \cite{zbontar2016stereo,Chen_2015_ICCV,luo2016efficient}, deep networks at first replaced single steps in the pipeline \cite{scharstein2002taxonomy}, moving then to optimization \cite{seki2016}, disparity selection \cite{Shaked_2017_CVPR} and refinement \cite{Gidaris_2017_CVPR}. 
The first end-to-end model was proposed by Mayer \etal{} \cite{Mayer_2016_CVPR}, deploying a 1D correlation layer to encode pixel similarities and feed them to a 2D network. In alternative, Kendall \etal{} \cite{Kendall_2017_ICCV} stacked features to build a cost volume, processed by 3D convolutions to obtain disparity values through a differentiable \textit{argmin} operation. These two pioneering works paved the way for more complex and effective 2D   \cite{Pang_2017_ICCV_Workshops,Liang_2018_CVPR,ilg2018occlusions} and 3D \cite{yu2018deep,Chang_2018_CVPR,Zhang2019GANet} architectures. Finally, multi-task frameworks combining stereo with semantic segmentation \cite{yang2018segstereo,dovesi2019real} and edge detection \cite{song2018edgestereo,song2019edgestereo} proved to be effective as well.
On the other hand, deep learning stereo methods able to learn directly from images largely alleviate the need for \gt{} labels. These have been used either for domain adaptation or for training from scratch a deep stereo network. In the former case, Tonioni \etal{} \cite{Tonioni_2017_ICCV,tonioni2019unsupervised} leveraged traditional algorithms and confidence measures, in \cite{Tonioni_2019_CVPR} developed a modular architecture able to be updated in real time leveraging image reprojection and in \cite{Tonioni_2019_learn2adapt} made use of meta-learning for the same purpose.
In the latter, an iterative schedule to train an unsupervised stereo CNN has been proposed in \cite{Zhou_2017_ICCV}, Godard \etal{} \cite{Godard_CVPR_2017} trained a naive stereo network using image reprojection. Zhong \etal{} first showed the fast convergence of 3D networks when trained with image reprojection \cite{ZhongArxiv2017}, then adopted a RNN LSTM network using stereo video sequences \cite{Zhong_ECCV_2018}.  
Wang \etal{} \cite{wang2019unos} improved their stereo network thanks to a rigid-aware direct visual odometry module, while in \cite{lai19cvpr} the authors exploited the relationship between optical flow and stereo. Joung \etal{} \cite{joung2019unsupervised} trained a network from scratch selecting good matches obtained by a pretrained model. Finally, in \cite{Smolyanskiy_2018_CVPR_Workshops} a semi-supervised framework leveraging raw LiDAR and image reprojection has been proposed. 

\textbf{Monocular depth.} Single image depth estimation is attractive, yet an \illpose{} problem. Nonetheless, modern deep learning strategies showed impressive performance up to a scale factor. 
The first successful attempt in this field followed a supervised paradigm \cite{Laina_3DV_2016,Liu_IEEE_2016,Eigen_2014}. Seminal works switching to self-supervision are \cite{Godard_CVPR_2017} and \cite{Zhou_2017_CVPR}, respectively requiring stereo pairs and monocular video sequences in place of \gt{} depth labels. Both methods paved the way for self-supervised monocular methods \cite{Poggi_3DV_2018,Godard_ICCV_2019}. In recent works \cite{Tosi_CVPR_2019,Watson_ICCV_2019}, \textit{proxies} labels have been distilled from traditional stereo algorithms \cite{SGM_PAMI} in order to strengthen the supervision from stereo pairs.

In parallel to our work, Watson \etal{} \cite{Watson_ECCV_2020} used monocular depth networks to train stereo models from single images through view synthesis.

\textbf{Depth completion.} Finally, we mention methods that aim at filling a sparse or low resolution depth map, traditionally output of a LiDAR, to obtain dense estimates. Two main categories exist, respectively based on depth only \cite{liu2015depth,ku2018defense,eldesokey2018propagating} or guided by images \cite{yang2007spatial,huang2019hms,Uhrig2017THREEDV,cheng2018depth,ma2019self,chen2019learning}. 
Although inspired by these works, our strategy is not comparable with them since processing very different input cues and deployed for other purposes.

\section{Method}

\begin{figure}[t]
    \centering
    \includegraphics[width=1.02\textwidth]{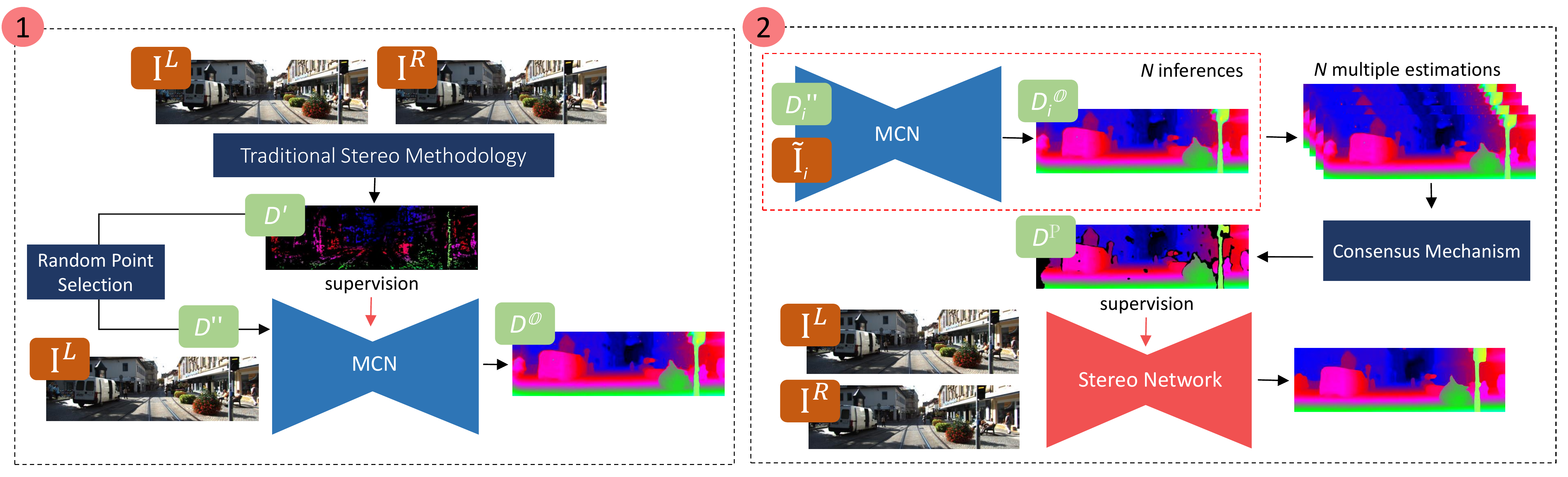}
     \caption{\textbf{Overview of our methodology.} \raisebox{.5pt}{\textcircled{\raisebox{-.9pt} {1}}}
 Sparse disparity points from a traditional stereo method are given as input to a monocular completion network (MCN). Then, in \raisebox{.5pt}{\textcircled{\raisebox{-.9pt} {2}}}
 we leverage MCN to distill accurate proxies through the proposed consensus mechanism. Such labels guide the training of a deep stereo network.}
    \label{fig:method}
\end{figure}

This section describes in detail our strategy, that allows us to distill highly accurate disparity annotations for a stereo dataset made up of raw rectified images only and then use them to supervise deep stereo networks. It is worth noting that, by abuse of notation, we use depth and disparity interchangeably although our proxy extraction method works entirely in the disparity domain. 
For our purposes, we rely on two main stages, as depicted in Fig. \ref{fig:method}: 1) we train a monocular completion network (MCN) from sparse disparity points sourced by traditional stereo methods and 2) we train deep stereo networks using highly reliable points from MCN, selected by a novel consensus mechanism. 

\subsection{Monocular Completion Network (MCN)}
Stereo algorithms struggle on occluded regions due to the difficulties to find correspondences between images. On the contrary, monocular methods do not rely on matching and thus, they are potentially not affected by this problem. In this stage, our goal is to obtain a strong guidance even on occluded areas relying on a monocular depth network. However, monocular estimates intrinsically suffer the scale factor ambiguity due to the lack of geometric constraints. Therefore, since stereo pairs are always available in our setup, we also leverage on reliable sparse disparity input points from traditional stereo algorithms in addition to the reference image. Thanks to this combination, MCN is able to predict dense depth maps preserving geometrical information.

\begin{figure}[t]
    \centering
    \begin{tabular}{c}
        \includegraphics[width=0.33\textwidth]{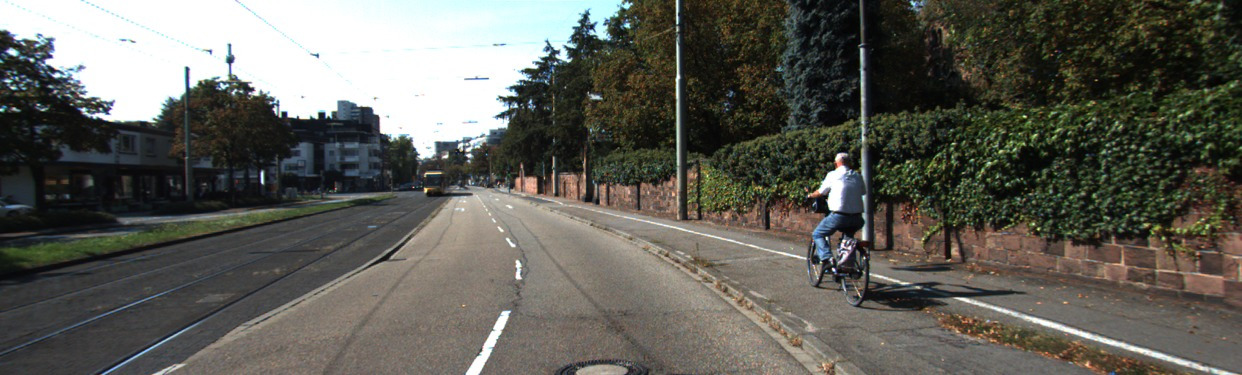} 
        \includegraphics[width=0.33\textwidth]{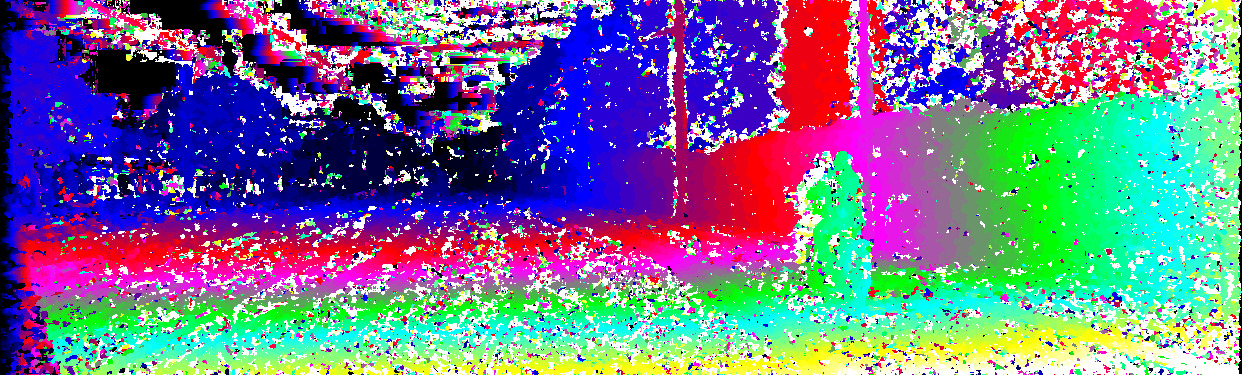} 
        \includegraphics[width=0.33\textwidth]{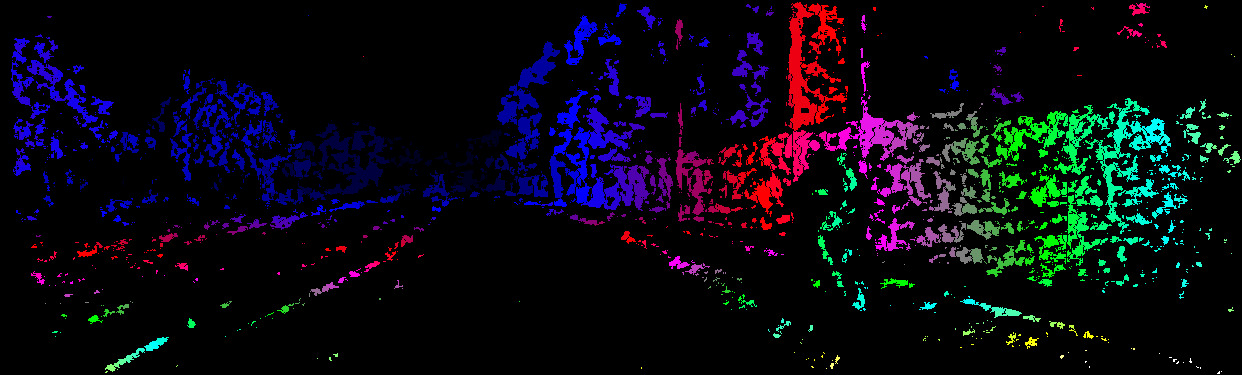}\\
    \end{tabular}
    \caption{\textbf{Disparity map filtering.} From left to right, reference image from KITTI, the noisy disparity map computed by \cite{Secaucus_1994_ECCV} and the outcome of filtering \cite{BMVC_2017}.}
    \label{fig:point_filtering}
\end{figure}

\textbf{Reliable disparity points extraction.}
At first, we rely on a traditional stereo matcher $\stereo$ (\eg{} \cite{Secaucus_1994_ECCV}) to obtain an initial disparity map $\dispL$ from a given stereo pair $(\imageL,\imageR)$ as 
\begin{equation}
    \dispL = \stereo(\imageL,\imageR)
\end{equation}
However, since such raw disparity map contains several outliers, especially on ill-posed regions such as occlusions or texture-less areas as it can be noticed in Fig. \ref{fig:point_filtering}, a filtering strategy $\filter$ (\eg{} \cite{BMVC_2017}) is applied to discard spurious points

\begin{equation}
    \dispL'=\filter( \stereo(\imageL,\imageR) )
\end{equation}
By doing so, only a subset $\dispL'$ of highly reliable points is preserved from $\dispL$ at the cost of a sparser disparity map. However, most of them do not belong to occluded regions thus not enabling supervision on such areas. This can be clearly perceived observing the outcome of a filtering strategy in Fig. \ref{fig:point_filtering}.

\begin{figure}[t]
    \centering
    \begin{tabular}{c}
        \includegraphics[width=0.33\textwidth]{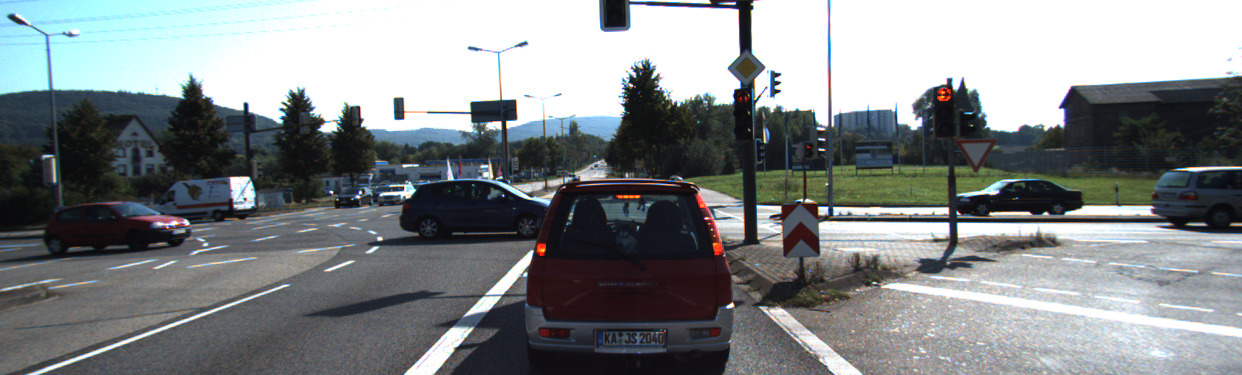} 
        \includegraphics[width=0.33\textwidth]{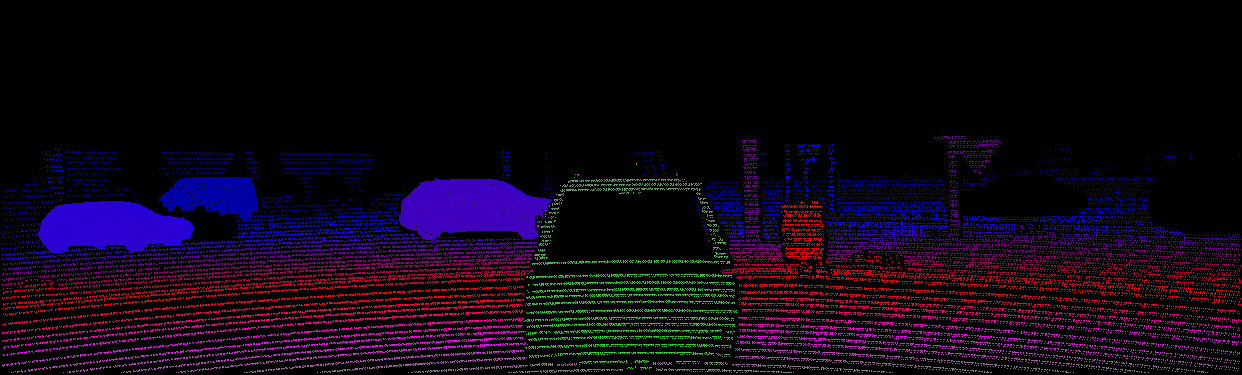} 
        \includegraphics[width=0.33\textwidth]{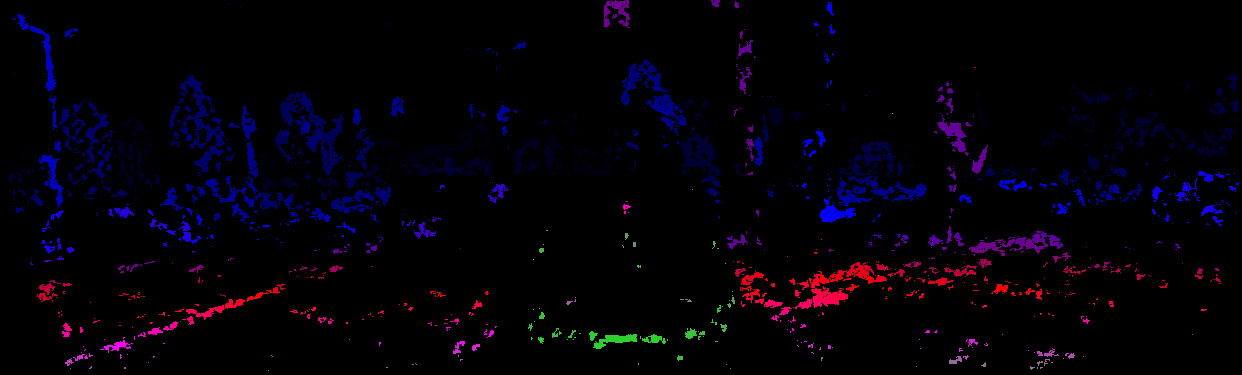} \\
        \includegraphics[width=0.33\textwidth]{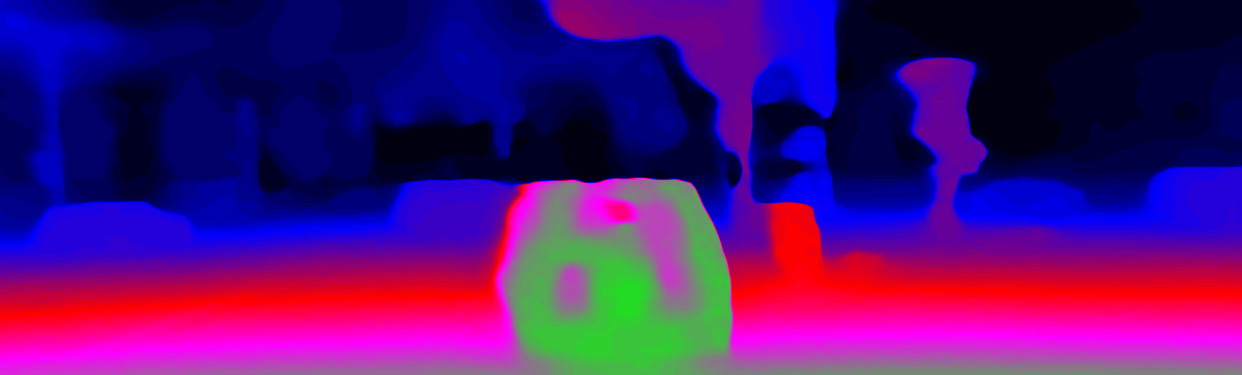} 
        \includegraphics[width=0.33\textwidth]{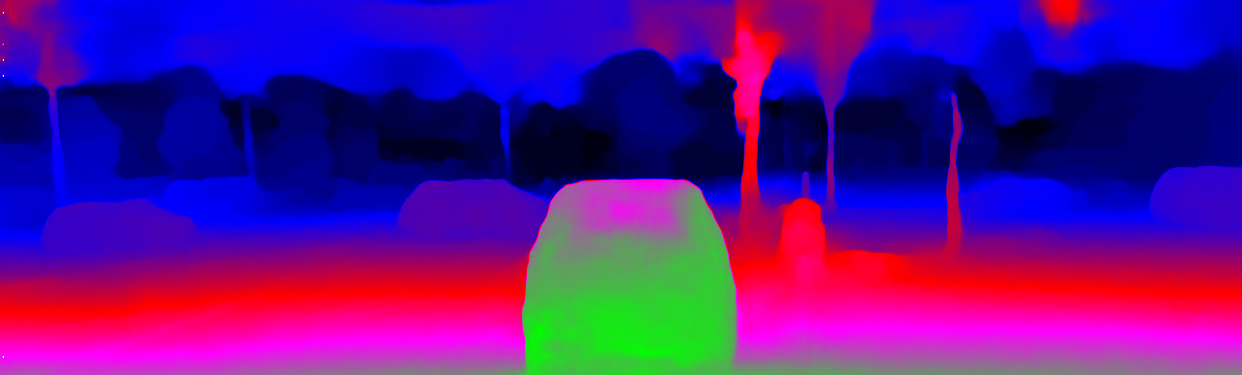}
        \includegraphics[width=0.33\textwidth]{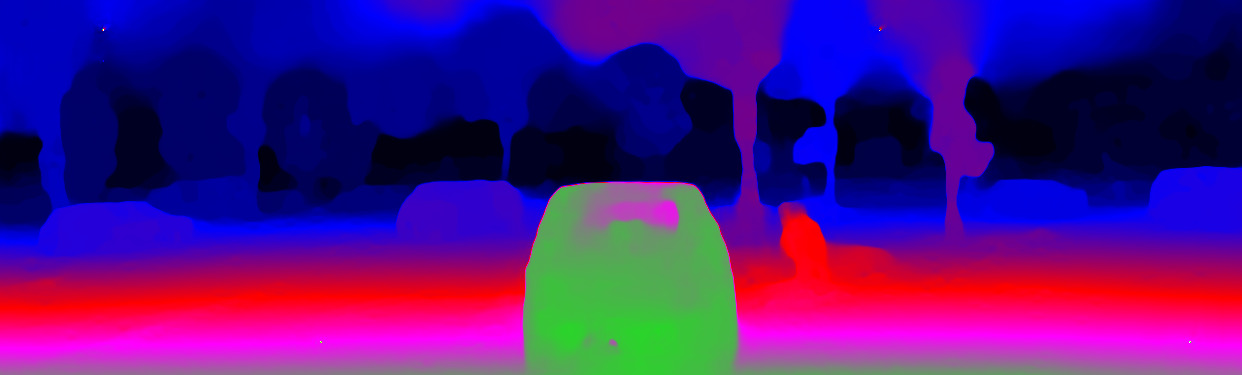} \\
        
        \includegraphics[width=0.33\textwidth]{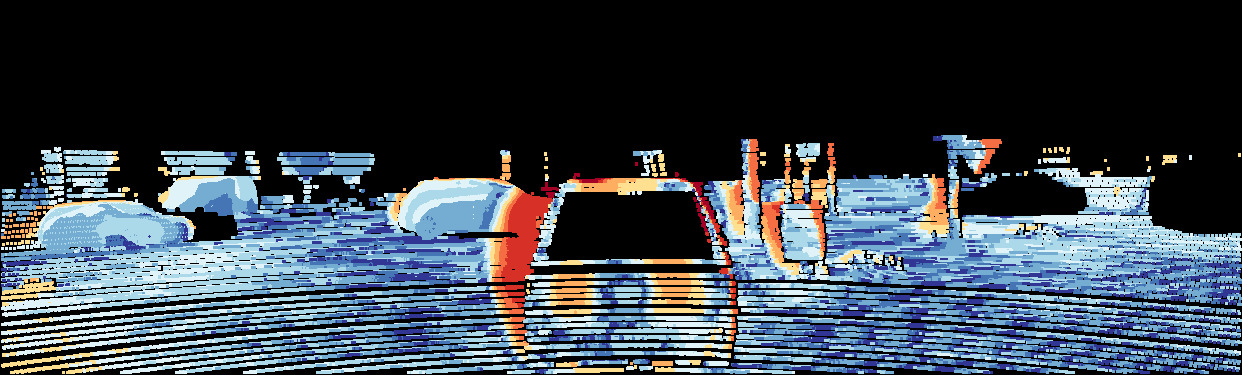}
        \includegraphics[width=0.33\textwidth]{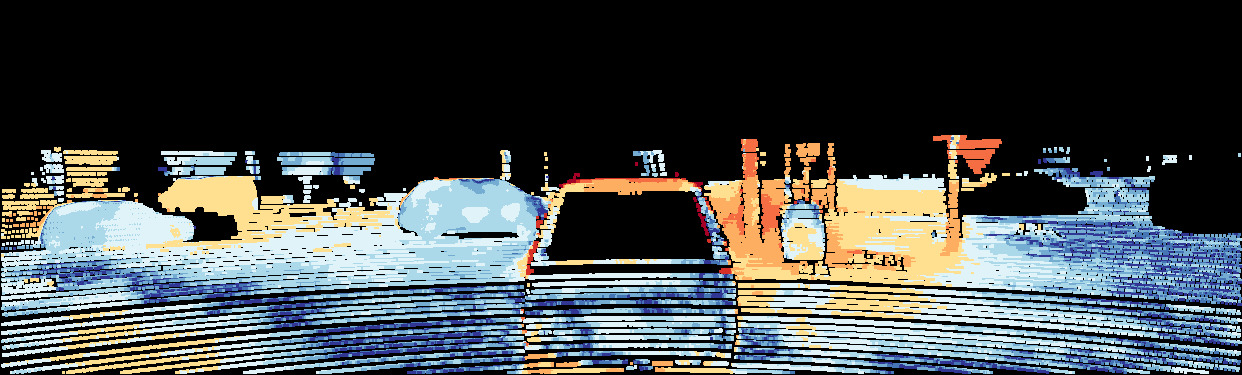} 
        \includegraphics[width=0.33\textwidth]{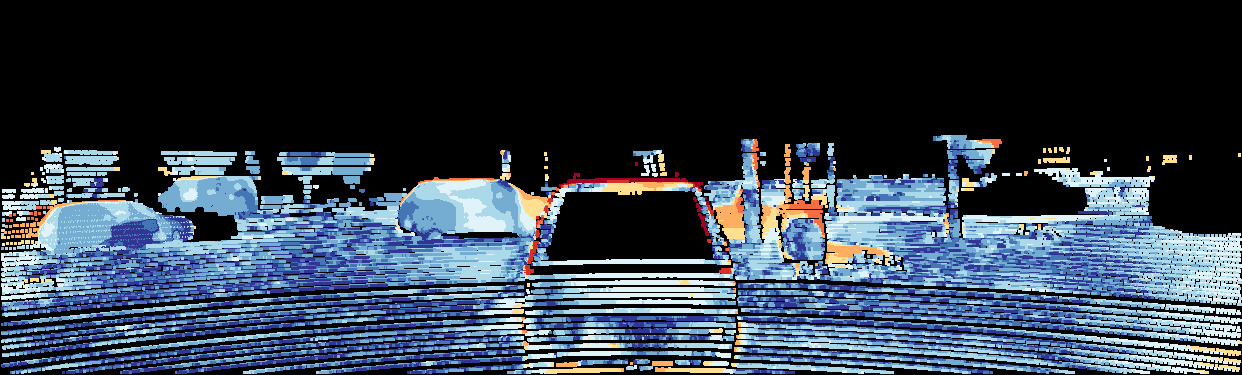} \\
        \includegraphics[width=\textwidth]{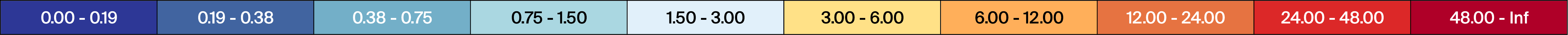}
    \end{tabular}
    \caption{\textbf{Occlusion handling and scale recovery.} The first row depicts the reference image from KITTI, the \gt{} and the disparity map by \cite{Secaucus_1994_ECCV} filtered with \cite{BMVC_2017}.
    In the middle, from left to right the output of monocular depth network \cite{Tosi_CVPR_2019} trained without occlusion augmentation, the same  network using the occlusion augmentation and our MCN. Last row shows the corresponding error maps. Best viewed with colors.}
    \label{fig:occlusion_handling}
\end{figure}

\textbf{Monocular Disparity Completion.}
Given $\dispL'$, we deploy a monocular completion network, namely MCN, in order to obtain a dense map $\dispD$. 
We self-supervise MCN from stereo and, as in \cite{Godard_ICCV_2019}, to handle occlusions we horizontally flip $(\imageL,\imageR)$ at training time with a certain probability without switching them. Consequently, occluded regions (\eg{} the left border of objects) are randomly swapped with not-occluded areas (\eg{} the right borders), preventing to always expect high error on left and low error on right borders, thus forcing the network to handle both. This strategy turns out ineffective in case of self-supervised stereo, since after horizontal flip the stereo pair have to be switched in order to keep the same search direction along the epipolar line, thus making occlusions occur in the same regions (see the supplementary material for details).
Even if this technique helps to alleviate errors in occluded regions, a pure monocular network struggles compared to a stereo method at determining the correct depth.
This is well-known in the literature and shown in our experiments as well.
Thus, we adopt a completion approach leveraging sparse reliable points provided by a traditional stereo method constraining the predictions to be properly scaled.
Given the set of filtered points, only a small subset $\dispL''$, with $||\dispL''|| \ll ||\dispL'||$, is randomly selected and used as input, while $\dispL'$ itself is used for supervision purposes. The output of MCN is defined as

\begin{equation}\label{eq:sampling}
    \dispD = \text{MCN}(\imageL, \dispL'' \xleftarrow{p \ll 1} (\dispL') )
\end{equation}
with $x \xleftarrow{p} (y)$ a random uniform sampling function extracting $x$ values out of $y$ per-pixel values with probability $p$. This sampling is crucial to both improve MCN accuracy, as shown in our experiments, as well as for the final distillation step discussed in the remainder.
Once trained, MCN is able to infer scaled dense disparity maps $\dispD$, as can be perceived in Fig. \ref{fig:occlusion_handling}. Looking at the rightmost and central disparity maps, we can notice how the augmentation protocol enables to alleviate occlusion artifacts. Moreover, our overall completion strategy, compared to the output of the monocular network without disparity seeds (leftmost and center disparity maps), achieves much higher accuracy as well as correctly handles occlusions.
Therefore, we effectively combine stereo from non-occluded regions and monocular prediction in occluded areas. Finally, we point out that we aim at specializing MCN on the training set to generate labels on it since its purpose is limited to distillation.

\subsection{Proxy Distillation for Deep Stereo} 
Eventually, we leverage the trained MCN to distill offline proxy labels beneficial to supervise stereo networks. However, such data might still contain some inconsistent predictions, as can be perceived in the rightmost disparity map of Fig. \ref{fig:occlusion_handling}. Therefore, our goal is to discard them, keeping trustworthy reliable depth estimates to train deep stereo networks.

\textbf{Consensus mechanism and distillation.} To this aim, given an RGB image $\mathcal{I}$ and the relative $\dispL'$, we perform $N$ inferences of MCN by feeding it with $\dispL''_i$ and $\tilde{\mathcal{I}}_i$, with $i\in[1,N]$. Respectively, $\dispL''_i$ is sampled from $\dispL'$ according to the strategy introduced in Sec. 3.1 and $\tilde{\mathcal{I}}_i$ is obtained through random augmentation (explained later) applied to $\mathcal{I}$.
This way, we exploit consistencies and contradictions among multiple $\dispD_i$ to obtain reliable proxy labels $\dispP$, defined as

\begin{equation}\label{eq:consensus}
    \dispP \xleftarrow{\sigma^2(\{\dispD_i\}^N_{i=1}) < \gamma} \mu(\{\dispD_i\}^N_{i=1})
\end{equation}
where $x \xleftarrow{\sigma^2(y)<\gamma} \mu(y)$ is a function that, given $N$ values $y$ for the same pixel, samples the mean value $\mu(y)$ only if the variance $\sigma^2(y)$ is smaller $\gamma$. Being distillation performed offline, this step does not need to be differentiable.

\begin{figure}[t]
    \centering
    \begin{tabular}{c}
        \includegraphics[width=0.33\textwidth]{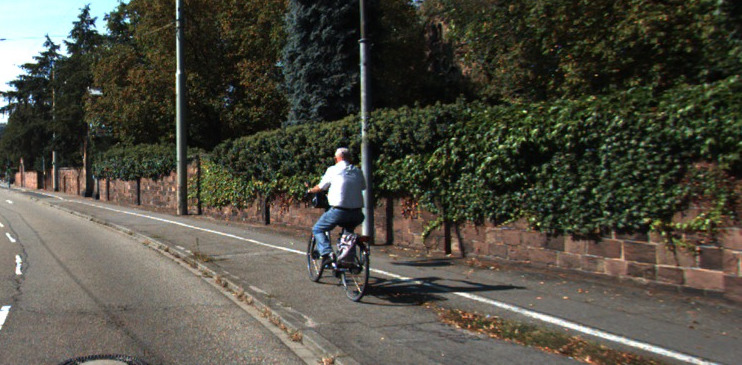}
        \includegraphics[width=0.33\textwidth]{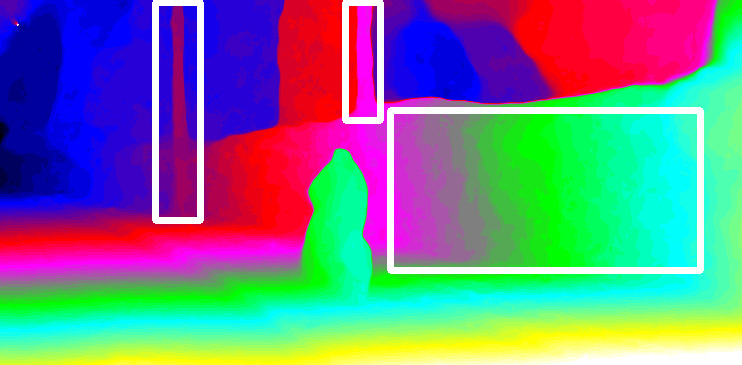}
        \includegraphics[width=0.33\textwidth]{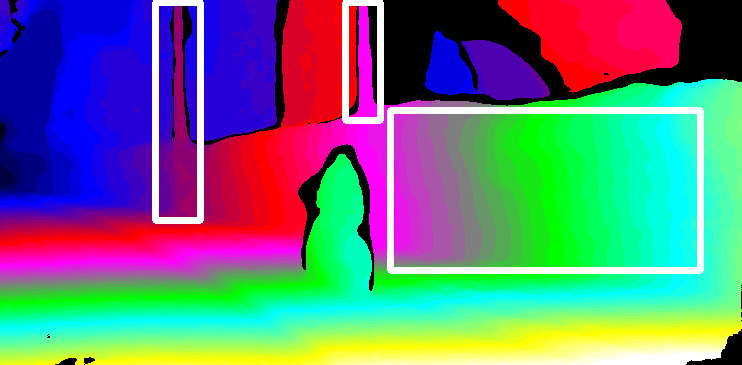} \\
        \includegraphics[width=0.33\textwidth]{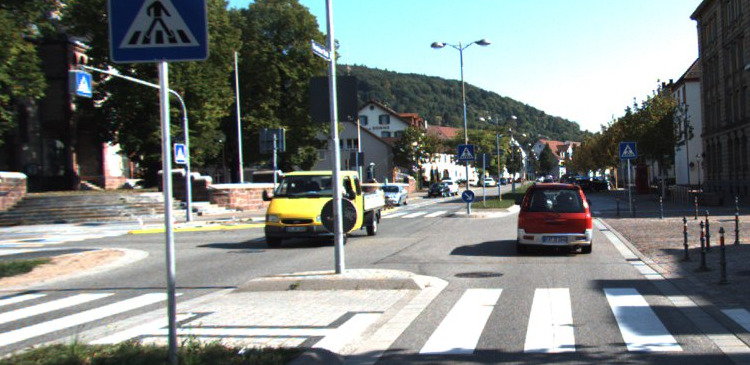}
        \includegraphics[width=0.33\textwidth]{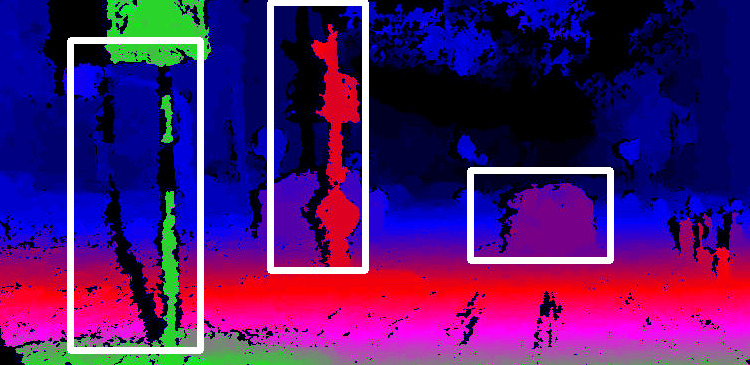}
        \includegraphics[width=0.33\textwidth]{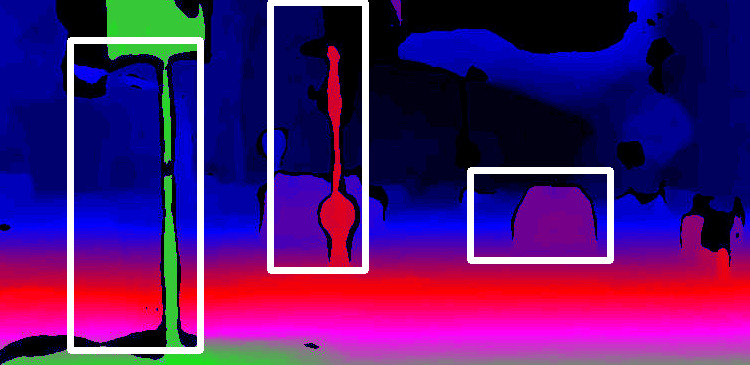}
    \end{tabular}
    \caption{\textbf{Proxy distillation.} The first row depicts, from left to right, the reference image, the disparity map computed by a single inference of MCN and the one filtered and regularized using our consensus mechanism. The second row shows the reference image, the disparity map generated by SGM \cite{SGM_PAMI} filtered using the left-right consistency check strategy and our disparity map. Images from KITTI.} 
    \label{fig:proxy_distillation}
\end{figure}

Fig. \ref{fig:proxy_distillation} shows that such a strategy allows us to largely regularize $\dispP$ compared to $\dispD$, preserving thin structures, \eg{} the poles on the right side, yet achieving high density. It also infers significant portions of occluded regions compared to proxies sourced from traditional methods (\eg{} SGM). 

\textbf{Deep Stereo Training.} Once highly accurate proxy labels $\dispP$ are available on the same training set, we exploit them to train deep stereo networks in a self-supervised manner. In particular, a regression loss is used to minimize the difference between stereo predictions and $\dispP$.

\section{Experiments}
In this section, we first introduce the datasets used in this work, then we thoroughly evaluate our proposal, proving that sourcing labels with a monocular completion approach is beneficial to train deep stereo networks. 

\subsection{Datasets}

\hspace*{5mm}\textbf{KITTI.} The KITTI (K) dataset \cite{Menze2015CVPR} contains 61 scenes (about 42,382 stereo pairs), with a typical image size of $1242 \times 375$, captured using a stereo rig mounted on a moving car equipped with a LiDAR sensor. We conducted experiments using all of the raw KITTI images for training excluding scenes from the KITTI 2015 training set containing 200 \gt{} images used for testing purposes. This results in a training split containing 29K rectified stereo images.

\textbf{DrivingStereo.} DrivingStereo \cite{yang2019drivingstereo} (DS) is a recent large-scale dataset depicting autonomous driving scenarios in various weather conditions, containing more than 180k stereo pairs with high-quality disparity annotations generated by means of a model-guided filtering method from multi-frame LiDAR points. For our purposes, we split the dataset into a training set and a testing set consisting of 97681 and 1k images respectively. 

\textbf{Middlebury v3.} The Middlebury v3 dataset \cite{scharstein2014high} provides 15 stereo pairs depicting indoor scenes, with high precision and dense \gt{} disparities obtained using structured light. We rely on this dataset for generalization purposes, using images and the ground-truth disparity maps at quarter resolution.

\textbf{ETH3D.} The ETH3D high-resolution dataset \cite{schops2017multi} depicts
heterogeneous scenes consisting of 27 grayscale stereo pairs with \gt{} depth values. As for Middlebury v3, we run generalization experiments on it.

\subsection{Implementation Details}

\hspace*{5mm}\textbf{Traditional stereo methods.} We consider two main non-learning based solutions, characterized by different peculiarities, to generate accurate sparse disparity points from a rectified stereo pair. In particular, we use the popular semi-global matching algorithm SGM \cite{SGM_PAMI}, exploiting the left-right consistency check (LRC) to remove wrong disparity assignments, and the WILD strategy proposed in \cite{BMVC_2017} that selects highly reliable values from the maps computed by the local algorithm Block-Matching (BM) \cite{Secaucus_1994_ECCV} exploiting traditional confidence measures. We refer to these methods (\ie{} stereo method followed by a filtering strategy) as SGM/L and BM/W, respectively.

\textbf{Monocular Completion Network.}
We adopt the publicly available self-supervised monocular architecture monoResMatch \cite{Tosi_CVPR_2019} trained with the supervision of disparity proxy labels specifically suited for our purposes. We modify the network to exploit accurate sparse annotations as input by concatenating them with the RGB image. We set the random sampling probability in Eq. \ref{eq:sampling} as $p=\frac{1}{1000}$.
In our experiments, we train from scratch the MCN network following the same training protocol defined in \cite{Tosi_CVPR_2019} except for the augmentation procedure which includes the flipping strategy (with $0.25$ probability) aimed at handling occlusion artifacts \cite{Godard_ICCV_2019}. Instead, we empirically found out that generating $\dispD$ using a larger set of points helps to achieve more accurate predictions at inference time. In particular, we fix $p=\frac{1}{20}$ and $p=\frac{1}{200}$ for BM/W and SGM/L respectively. Finally, for the consensus mechanism, we fix $N=50$, the threshold $\gamma=3$ and apply for each $\mathcal{I}_i$ color augmentation and random horizontal flip (with $0.5$ probability). Please see the supplementary material for more details about hyper-parameters. 

\textbf{Stereo networks.} We considered both 2D and 3D deep stereo architectures, ensuring a comprehensive validation of our proposal. In particular, we designed a baseline architecture, namely \stereodepth{}, by extending \cite{Godard_ICCV_2019} to process stacked left and right images, and \iresnet{} as examples of the former case, while \psm{} and GWCNet \cite{guo2019group} as 3D architectures.
At training time, the models predict disparities $\mathcal{D}^S$ at multiple scales in which each intermediate prediction is upsampled at the input resolution. A weighted smooth L1 loss function (the lower the scale, the lower the weight) minimizes the difference between $\mathcal{D}^S$ and the disparity provided by the proxy $\dispP$ considering only valid pixels, using Adam \cite{Kingma_2014} as optimizer ($\beta_1=0.9$ and $\beta_2= 0.999$). 
We adopt the original PyTorch \cite{PyTorch} implementation of the networks if available. Moreover, all the models have been trained to fit a single Titan X GPU. More details are provided in the supplementary material.

\subsection{Evaluation of Proxy Label Generators}\label{ablation:proxy_generation}
At first, we first evaluate the accuracy of proxies produced by our self-supervised approach with respect to traditional methods.
We consider both D1 and EPE, computed on disparities, as error metrics on both non-occluded (\textit{Noc}) and all regions (\textit{All}). In particular, D1 represents the percentage of pixels greater than 3 or more than 5\% of \gt{}, while EPE is obtained by averaging the absolute difference between predictions and \gt{} values. In addition, the density and the overlap with the \gt{} are reported to take into account filtering strategies. 
Table \ref{table:evaluation_proxy_generator} reports a thorough evaluation of different methodologies and filtering techniques. It can be noticed how BM and SGM have different performances due to their complementarity (local vs semi-global), but containing several errors. Filtering strategies help to remove outliers, at the cost of sparser maps. Notice that restoring the full density through \textit{hole-filling} \cite{SGM_PAMI} slightly improves the results of SGM/L, but it is not meaningful for BM/W since filtered maps are too sparse. Unsurprisingly, even if the depth maps produced by the vanilla monoResMatch are fully-dense, they are not accurate due to its inherent monocular nature. On the contrary, our monocular strategy MCN produces dense yet accurate maps thanks to the initial disparity guesses, regardless the sourcing stereo algorithm. Moreover, by applying Augmentation techniques (A) on the RGB image or selecting Random input points (R), allow to increase variance and to exploit our Consensus mechanism (C) to filter out unreliable values, thus achieving even better results. In fact, the consensus mechanism is able to discard wrong predictions preserving high density, reaching best performances when A and R are both applied. It is worth noting that if R is not performed, the network is fed with all the available guesses both at training and testing time, with remarkably worse results compared to configuration using random sampling.

%-------------------------------------------------------------------------
% KITTI 2015 test
%-------------------------------------------------------------------------

\begin{table*}[t!]
\centering
\resizebox{\columnwidth}{!}{

\begin{tabular}{l|ccccc|cc|cc|cc}

\hline

Method &  \multicolumn{5}{c}{Configuration} & \multicolumn{2}{c}{Statistics} & \multicolumn{2}{c}{Noc} & \multicolumn{2}{c}{All}\\
 &  Source & Filter & A & R & C & Density(\%) & Overlap(\%) & \cellcolor{lower} D1(\%) & \cellcolor{lower} EPE & \cellcolor{lower} D1(\%) & \cellcolor{lower} EPE \\

\hline
MONO & monoResMatch & - & -&  - & - & 100.0 & 100.0 & 26.63 & 2.96 &  27.00 & 2.99 \\

\hline
\hline
BM & BM & - & - & - & - &  100.0 & 100.0 & 34.48 & 16.14 & 35.46 & 16.41\\
SGM & SGM & - & - & - & - &  100.0 & 100.0 & 6.65 & 1.67 & 8.12 & 2.16\\ 

\hline
\hline
BM/L & BM & LRC &  - & - & - & 57.89 & 62.09 & 16.09 & 6.42 & 16.22 & 6.46\\ 
SGM/L & SGM & LRC &  - & - & - &86.47 & 92.28 & 3.99 & 1.00 & 4.01 & 1.00\\
SGM/L(\textit{hole-filling}) & SGM & LRC &  - & - & - &100.0 & 100.0 & 6.56 & 1.34 & 7.68 & 1.57\\
BM/W & BM & WILD &  - & - & - & 12.33 & 10.43 & 1.33 & 0.81 & 1.35 & 0.81\\ 

\hline
\hline
MCN-SGM/L  & SGM & LRC &  - & - &  - & 100.0 & 100.0 & 6.36 & 1.27 & 7.80 & 1.50\\ 
MCN-SGM/L-R  & SGM & LRC &  - & \checkmark &  - & 100.0 & 100.0 & 5.28 & 1.13 & 5.73 & 1.21\\ 
MCN-SGM/L-AC & SGM & LRC & \checkmark & - &  \checkmark & 95.36 & 97.36 & 5.58 & 1.17 & 5.58 & 1.15\\ 
MCN-SGM/L-RC & SGM & LRC & - & \checkmark & \checkmark &  93.50 & 96.32 & 2.95 & 0.86 & 3.14 & 0.89\\
MCN-SGM/L-ARC & SGM & LRC & \checkmark & \checkmark & \checkmark &  92.53 & 95.76 & 2.78 & 0.84 & 2.92 & 0.86\\
\hline
MCN-BM/W & BM & WILD & - & - & - & 100.0 & 100.0 & 11.86 & 1.93 & 12.50 & 2.03\\ 
MCN-BM/W-R & BM & WILD & - & \checkmark &  - & 100.0 & 100.0 & 6.79 & 1.40 & 7.11 & 1.45\\ 
MCN-BM/W-AC & BM & WILD & \checkmark & - &  \checkmark & 91.45 & 94.76 & 8.36 & 1.53 & 8.64 & 1.57\\ 
MCN-BM/W-RC & BM & WILD & - & \checkmark  & \checkmark & 91.12 & 95.28 & 3.79 & 0.95 & 4.03 & 1.0\\ 
MCN-BM/W-ARC & BM & WILD & \checkmark & \checkmark  & \checkmark & 86.82 & 93.56 & 3.16 & 0.90 & 3.27 & 0.92\\ 
\hline
\end{tabular}
}
\caption{\textbf{Evaluation of proxy generators.} We tested proxies generated by different strategies on the KITTI 2015 training set.} 
\label{table:evaluation_proxy_generator}
\vspace{-10pt}
\end{table*}

\textbf{Disparity completion comparison.} We validate our MCN combined with the consensus mechanism comparing it to GuideNet \cite{yang2019drivingstereo}, a supervised architecture designed to generate high-quality disparity annotations exploiting multi-frame LiDAR points and stereo pairs as input. Following \cite{yang2019drivingstereo}, we measure the valid pixels, correct pixels, and accuracy (\ie{} 100.0 - D1) on 142 images of the KITTI 2015 training set. Table \ref{table:guide_vs_ours} clearly shows how MCN trained in a self-supervised manner achieves comparable accuracy with respect to GuideNet-LiDAR by exploiting sparse disparity estimates from both \cite{BMVC_2017} and \cite{SGM_PAMI} but with a significantly higher number of points, even on foreground regions (Obj). Notice that LiDAR indicates that the network is fed with LiDAR points filtered according to \cite{Uhrig2017THREEDV}.
To further demonstrate the generalization capability of MCN to produce highly accurate proxies relying on points from heterogeneous sources, we feed MCN-BM/W-R with raw LiDAR measurements. By doing so, our network notably outperforms GuideNet in this configuration, despite it leverages a single RGB image and has not been trained on LiDAR points.

\subsection{Ablation study}
In this subsection, we support the statement that a completion approach provides a better supervision compared to traditional stereo algorithms. We first run experiments on KITTI and then use our best configuration on DrivingStereo as well, showing that it is effective on multiple large stereo datasets.     

\begin{table}[t]
\centering
\scalebox{0.9}{
\begin{tabular}{l|ccc|ccc}
\hline
Model & \multicolumn{3}{c}{All} & \multicolumn{3}{c}{Obj} \\
 & Valid & \cellcolor{upper} Correct & \cellcolor{upper} Accuracy (\%) & Valid & \cellcolor{upper}Correct & \cellcolor{upper} Accuracy (\%)\\
\hline
\textbf{MCN}-BM/W-ARC & 11,551,461 & 11,247,966 & 97.37 & 1,718,267 & 1,642,872 & 95.61 \\
\textbf{MCN}-SGM/L-ARC & 12,201,763 & 11,860,923 & 97.20 & 1,788,154 & 1,672,222 & 93.52 \\
\textbf{MCN}-LiDAR & 11,773,897 & 11,636,787 & 98.83 & 1,507,222 & 1,459,726 & 96.84  \\
\hline
GuideNet-LiDAR \cite{yang2019drivingstereo} \textdagger & 2,973,882 & 2,915,110 & 98.02 & 221,828 & 210,912 & 95.07 \\
\hline
\end{tabular}
}
\caption{\textbf{Model-guided comparison}. Comparison between our self-supervised MCN model and the  supervised GuideNet stereo architecture \cite{yang2019drivingstereo} using 142 \gt{} images of the KITTI 2015 training set. \textdagger{} indicates that the network requires LiDAR points at training time. Accuracy is defined as $100$-D1.}
\label{table:guide_vs_ours}
\end{table}

%-------------------------------------------------------------------------
% KITTI 2015 test
%-------------------------------------------------------------------------

\begin{table}[t]
\centering
\setlength{\tabcolsep}{8pt} %% default is 6pt
\scalebox{0.8}{
\begin{tabular}{lc|cc|cc}
\hline
Backbone & Supervision & \multicolumn{2}{c}{Noc} & \multicolumn{2}{c}{All}\\
 & & \cellcolor{lower}D1(\%) & \cellcolor{lower}EPE & \cellcolor{lower}D1(\%) & \cellcolor{lower}EPE \\
\hline
\stereodepth  & PHOTO & \textbf{6.50} & \textbf{1.30} & \textbf{7.12} & \textbf{1.40} \\ 
PSMNet & PHOTO & 6.62 & \textbf{1.30} & 7.67 & 1.50 \\
\hline
\stereodepth  & SGM-L & 5.22 & \textbf{1.13} & 5.43 & \textbf{1.15} \\ 
\stereodepth  & SGM/L(\textit{hole-filling}) & 6.06 & 1.16 & 6.38 & 1.21 \\ 
\stereodepth  & BM/W & \textbf{5.19} & 1.16 & \textbf{5.37} & 1.18 \\ 

PSMNet  & SGM/L & 5.46 & 1.19 & 5.61 & 1.21 \\
PSMNet  & SGM/L(\textit{hole-filling}) & 6.06 & 1.23 & 6.32 & 1.26 \\ 
PSMNet & BM/W & 6.89 & 1.59 & 7.03 & 1.60 \\ 

\hline
\stereodepth  & MCN-SGM/L-R & 5.11 & 1.11 & 5.37 & 1.14 \\ 
\stereodepth & MCN-BM/W-R & 4.75 & 1.05 & 4.96 & 1.07 \\
\stereodepth  & MCN-SGM/L-ARC & 4.56 & 1.08 & 4.77 & 1.11 \\ 
\stereodepth & MCN-BM/W-ARC & 4.21 & 1.06 & 4.39 & 1.07 \\

PSMNet & MCN-SGM/L-R & 4.39 & 1.05 & 4.60 & 1.07 \\ 
PSMNet & MCN-BM/W-R & 4.30 &  1.06 & 4.49 & 1.08 \\
PSMNet  & MCN-SGM/L-ARC & 4.02 & 1.05 & 4.20 & 1.07 \\ 
PSMNet  & MCN-BM/W-ARC & \underline{\textbf{3.68}} & \underline{\textbf{0.99}} & \underline{\textbf{3.85}} & \underline{\textbf{1.01}} \\
\hline
\stereodepth  & LiDAR/SGM \cite{Uhrig2017THREEDV} & 3.95 & 1.07 & 4.10 & 1.09 \\ 
PSMNet  & LiDAR/SGM \cite{Uhrig2017THREEDV} & \textbf{3.93} & \textbf{1.05} & \textbf{4.07} & \textbf{1.07} \\
\hline
\end{tabular}
}
\caption{\textbf{Ablation study.} We trained Stereodepth and PSMNet on KITTI using supervision signals from different proxy generators and tested on KITTI 2015.}
\label{table:ablation}
\end{table}

\textbf{KITTI.} For the ablation study, reported in Table \ref{table:ablation}, we consider both 3D (PSMNet) and 2D (\stereodepth{}) networks featuring different computational complexity.
First, we train the baseline configuration of the networks, \ie{} relying image reconstruction loss functions (PHOTO) only as in \cite{Godard_ICCV_2019} (see supplementary material for more details). Then, we leverage disparity values sourced by traditional stereo algorithms in which outliers have been removed by the filtering strategies adopted. Such labels provide a useful guidance for stereo networks and allow to obtain more accurate models w.r.t. the baselines. Nonetheless, proxies produced by MCN prove to be much more effective than traditional ones, improving both D1 and EPE by a notable margin regardless the stereo algorithm used to extract the input guesses. Moreover, it can be perceived that best results are obtained when the complete consensus mechanism is enabled.

\begin{figure}[t]
    \centering
    \begin{tabular}{c}
        \includegraphics[width=0.28\textwidth]{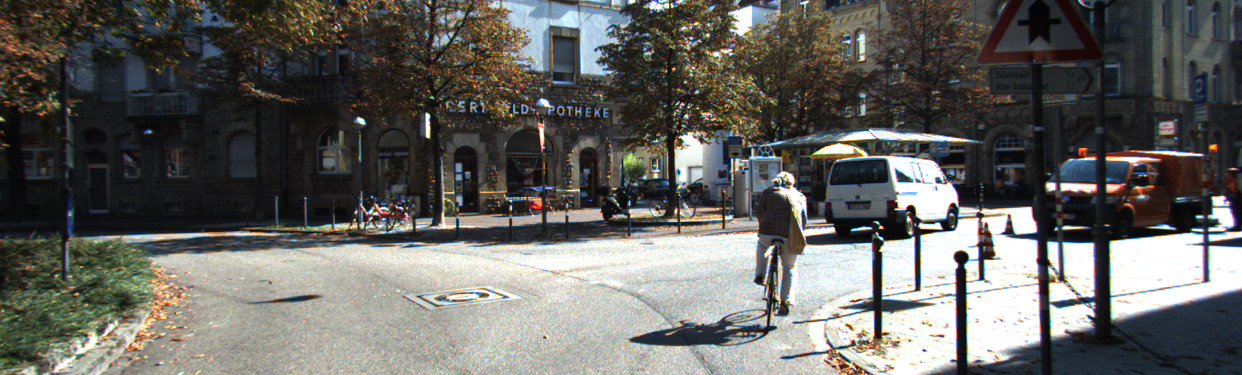} 
        \includegraphics[width=0.28\textwidth]{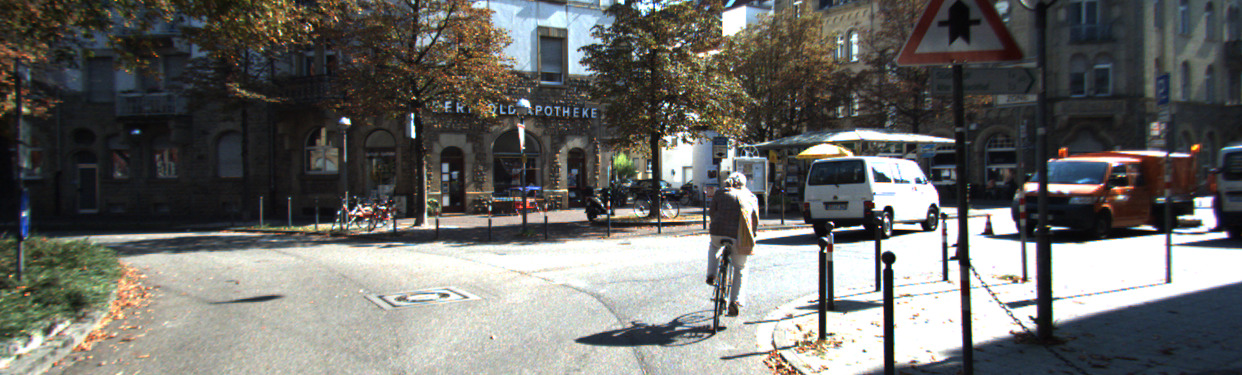} 
        \includegraphics[width=0.28\textwidth]{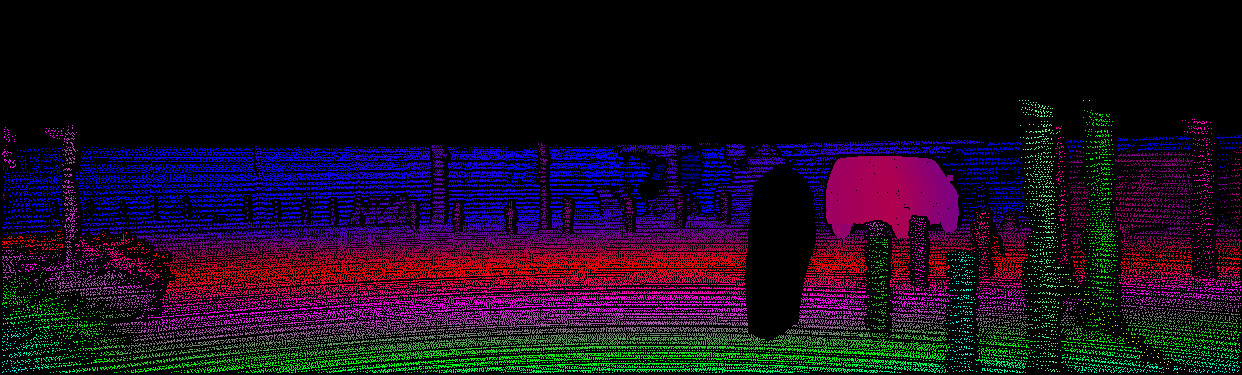} \\
        \includegraphics[width=0.28\textwidth]{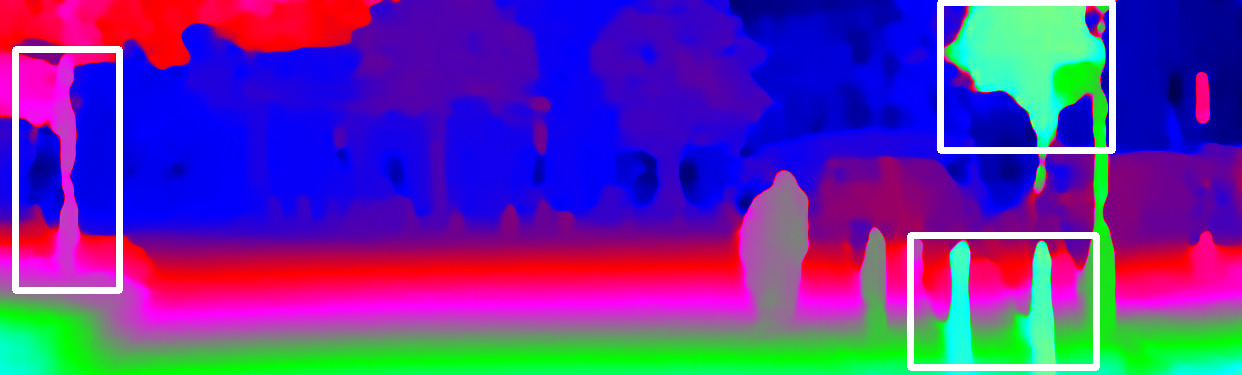} 
        \includegraphics[width=0.28\textwidth]{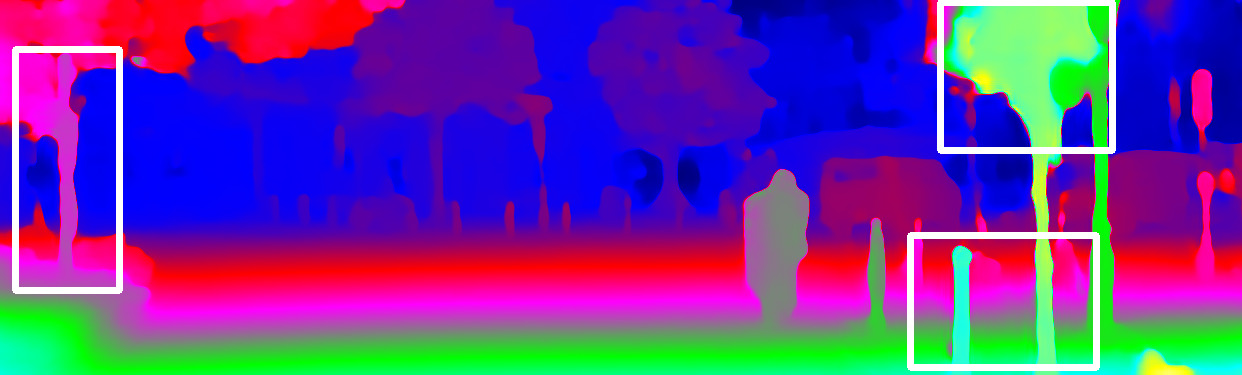}
        \includegraphics[width=0.28\textwidth]{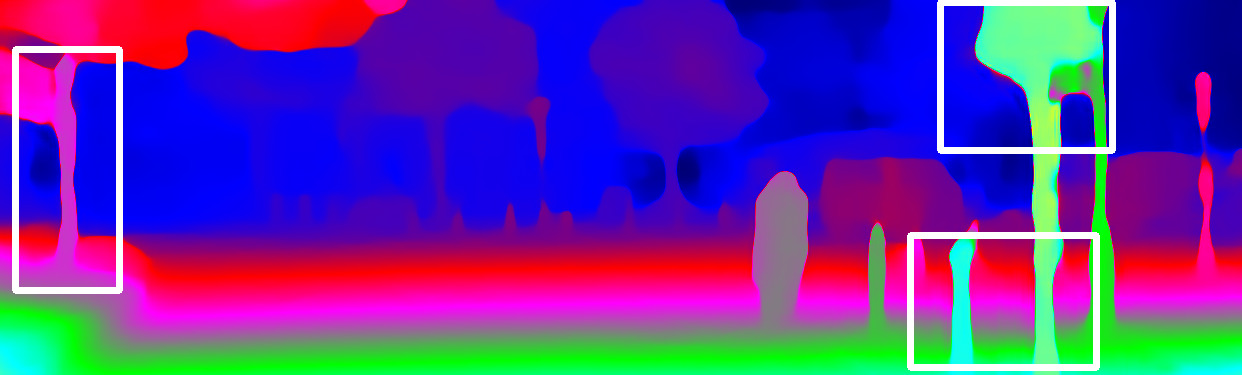} \\
        \includegraphics[width=0.28\textwidth]{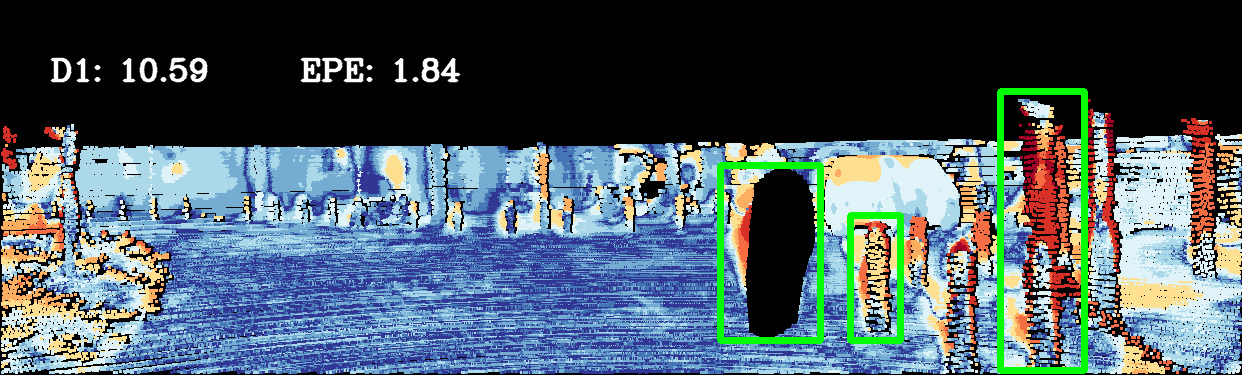} 
        \includegraphics[width=0.28\textwidth]{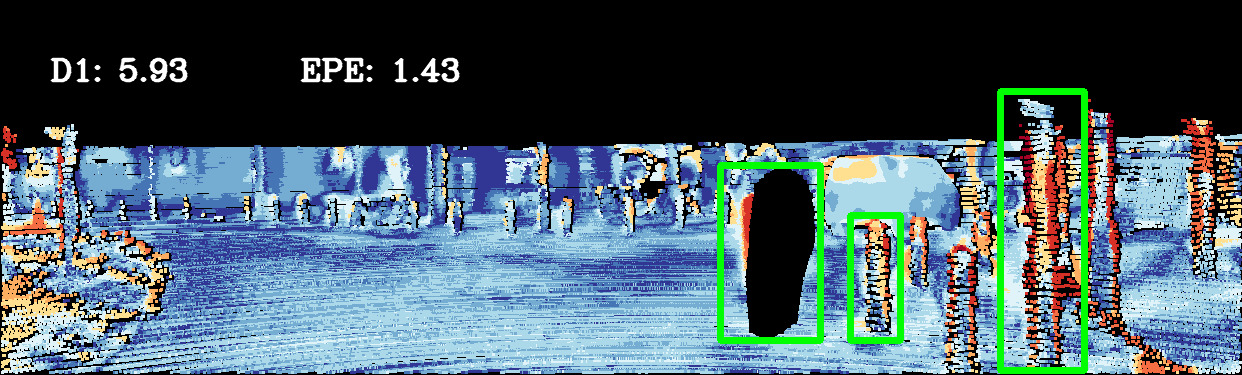}
        \includegraphics[width=0.28\textwidth]{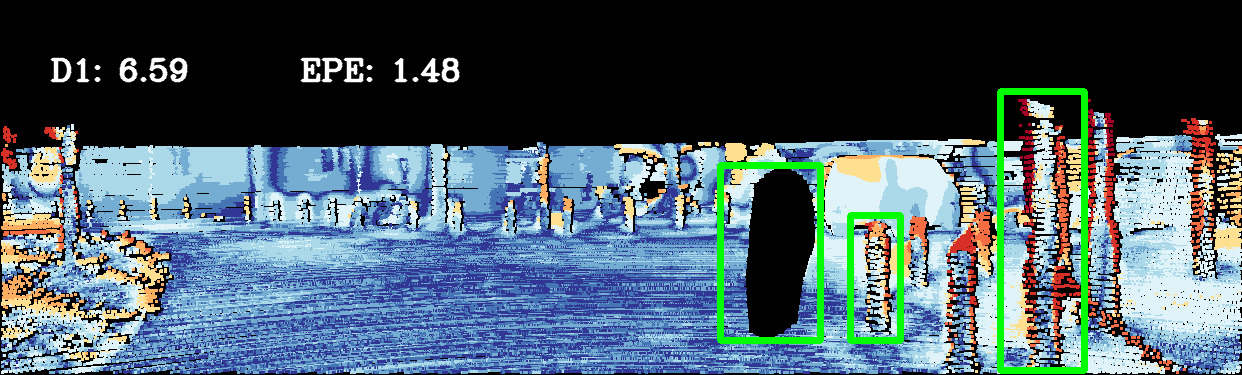} \\
        \includegraphics[width=0.86\textwidth]{images/ablation//kitti_scale-eps-converted-to.pdf}
    \end{tabular}
    \caption{\textbf{Impact of proxies.} From top, input stereo pair and \gt{} disparity map, predictions by \stereodepth{} trained with SGM/L (left), LiDAR (center) and our MCN-BM/W-ARC (right), error maps. Best viewed with colors.}
    \label{fig:method_comparison}
\end{figure}

Finally, we rely also on filtered LiDAR measurements from \cite{Uhrig2017THREEDV} in order to show differences with respect to supervision from active sensors.
Noteworthy, models trained using proxies distilled by ARC configuration of MCN prove to be comparable or even better than using LiDAR with PSMNet and \stereodepth{}. This behaviour can be explained due to a more representative and accurate supervision on occluded areas than traditional stereo and filtered LiDAR, thus making the deep networks more robust even there, as clearly shown in Fig. \ref{fig:method_comparison}.

\textbf{DrivingStereo.} We validate the proposed strategy also on DrivingStereo, proving that our distillation approach is able to largely improve the performances of stereo networks also on different datasets. In particular, in Table \ref{table:ds} we compare \stereodepth{} and PSMNet errors when trained using MCN-BM/W-ARC method (\ie{} the best configuration on KITTI) with LiDAR and BM/W. Again, our proposal outperforms BM/W, and reduces the gap with high quality LiDAR supervision.
Moreover, to verify generalization capabilities, we test on KITTI also correspondent models trained on DrivingStereo, without performing any fine-tuning (DS $\rightarrow$ K), and vice versa (K $\rightarrow$ DS). It can be noticed that the gap between KITTI models (see Table \ref{table:ablation}) and those trained on DrivingStereo gets smaller, proving that the networks are able to perform matching correctly even in cross-validation scenario. We want to point out that this is due to our proxies, as can be clearly perceived by looking at rows 1-2 vs 3-4 in Table \ref{table:ds}.

%-------------------------------------------------------------------------
% DrivingStereo experiments
%-------------------------------------------------------------------------

\begin{table}[t]
\centering
\setlength{\tabcolsep}{6pt} %% default is 6pt
\scalebox{0.9}{
\begin{tabular}{l|c|cc|cc|cc}
\multicolumn{2}{c}{}&\multicolumn{4}{c}{Source $\rightarrow$ Target}\\
\hline
Backbone & Supervision & \multicolumn{2}{c}{DS $\rightarrow$ DS} & \multicolumn{2}{c}{K $\rightarrow$ DS} & \multicolumn{2}{c}{DS $\rightarrow$ K}\\
 &  & \cellcolor{lower} D1(\%) & \cellcolor{lower} EPE & \cellcolor{lower} D1(\%) & \cellcolor{lower} EPE & \cellcolor{lower} D1(\%) & \cellcolor{lower} EPE\\
 
\hline
\stereodepth & BM/W & \textbf{4.46} & \textbf{1.20} & \textbf{4.67} & \textbf{1.10} & \textbf{6.35} & \textbf{1.36} \\
PSMNet & BM/W & 8.81 & 1.94 & 5.06 & 1.30 & 7.07 & 1.65\\ 
\hline
\stereodepth & MCN-BM/W-ARC & 2.47 & 0.94 & 2.97 & 0.96 & 5.64 & 1.22 \\
PSMNet & MCN-BM/W-ARC & \textbf{1.87} & \textbf{0.86} & \underline{\textbf{2.32}} & \underline{\textbf{0.88}} & \textbf{5.16} & \underline{\textbf{1.17}}  \\ 

\hline
\stereodepth & LiDAR \cite{yang2019drivingstereo} & 1.20 & 0.69 & 3.60 & 1.23 & 4.57 & \underline{\textbf{1.17}} \\
PSMNet  & LiDAR \cite{yang2019drivingstereo} & \underline{\textbf{0.59}} & \underline{\textbf{0.54}} & \textbf{2.64} & \textbf{1.03} & \underline{\textbf{4.52}} & 1.26 \\ 
\hline
\end{tabular}
}
\caption{\textbf{Cross-validation analysis.} We tested on the Target dataset models trained on the Source one, leveraging different proxies. Notice that no fine-tuning on the target dataset is performed in case of cross-validation.
}

\label{table:ds}
\end{table}

%-------------------------------------------------------------------------
% KITTI 2015 test
%-------------------------------------------------------------------------
\begin{table}[t]
\centering
\scalebox{0.9}{
\begin{tabular}{l|cccc|ccc}

\hline
Method & \cellcolor{lower} RMSE & \cellcolor{lower} RMSE log & \cellcolor{lower} D1 (\%) & \cellcolor{lower} EPE &  \cellcolor{upper}$\delta<$1.25 &  \cellcolor{upper}$\delta<1.25^2$ & \cellcolor{upper}$\delta<1.25^3$ \\
\hline

\hline
Godard et al.\cite{Godard_CVPR_2017} (stereo)  & 5.742 & 0.202 & 10.80 & - & 0.928 & 0.966 &  0.980\\
Lai et al.\cite{lai19cvpr} & 4.186 & 0.157 & 8.62 & 1.46 & 0.950 &  0.979 & 0.990\\
Wang et al.\cite{wang2019unos} (stereo only)  & 4.187 & 0.135 & 7.07 & - & 0.955 & 0.981 & 0.990\\
Zhong et al.\cite{ZhongArxiv2017} & 4.857 & 0.165 & 6.42 & - & 0.956 & 0.976 & 0.985\\
Wang et al.\cite{wang2019unos} (stereo videos) & 3.404 & 0.121 & 5.94 & - & 0.965 & 0.984 & 0.992 \\
Zhong et al.\cite{Zhong_ECCV_2018}* & \underline{\textbf{(3.176)}} & (0.125) & (5.14) & - & (0.967) & - & -\\

\textbf{Ours} (\stereodepth{}) & 3.882 &  0.117 &  4.39 & 1.07 & 0.971 &  0.988 &   \underline{\textbf{0.993}} \\
\textbf{Ours} (GWCNet) & 3.614 & 0.111 & 3.93 & 1.04 & 0.974 &  \underline{\textbf{0.989}} &  \underline{\textbf{0.993}} \\ 
\textbf{Ours} (IResNet)  & 3.464 &  \underline{\textbf{0.108}} & 3.88 & 1.02 &  \underline{\textbf{0.975}}  & 0.988 &   \underline{\textbf{0.993}}\\
\textbf{Ours} (PSMNet)  & 3.764 & 0.115 & \underline{\textbf{3.85}} &  \underline{\textbf{1.01}} & 0.974 & 0.988 &  \underline{\textbf{0.993}} \\
\hline
\end{tabular}
}
\caption{\textbf{Comparison with state-of-the-art}. Results of different self-supervised stereo networks on the KITTI 2015 training set with max depth set to 80m. \textbf{Ours} indicates networks trained using MCN-BM/W-ARC labels. * indicates networks trained on the same KITTI 2015 data, therefore not directly comparable with other methods.}
\label{table:comparison_sota}
\end{table}

\begin{table}[t]
\centering
\scalebox{0.9}{

\begin{tabular}{l|cc|cccc}
\hline
Models & Dataset & E2E & \cellcolor{lower} D1-bg (\%) & \cellcolor{lower} D1-fg (\%) & \cellcolor{lower} D1-All (\%) & \cellcolor{lower} D1-Noc (\%) \\\hline

Zbontar and LeCun (acrt) \cite{zbontar2016stereo} & \textcolor{red}{K} & -  & 2.89 & 8.88 & 3.89 & 3.33\\ 
Tonioni et al. \cite{Tonioni_2019_CVPR} &  \textcolor{red}{SF+K} & \checkmark & 3.75 & 9.20 & 4.66 & 4.27\\ 
Mayer et al. \cite{Mayer_2016_CVPR}  & \textcolor{red}{SF+K} & \checkmark &  4.32 & 4.41 & 4.34 & 4.05\\ 
Chang and Chen \cite{Chang_2018_CVPR} (PSMNet) & \textcolor{red}{SF+K}& \checkmark &  1.86 & 4.62 & 2.32 & 2.14 \\
Guo et al. \cite{guo2019group} (GWCNet) &  \textcolor{red}{SF+K} & \checkmark & 1.74 & 3.93 & 2.11 & 1.92 \\ 

Zhang et al. \cite{Zhang2019GANet}  &  \textcolor{red}{SF+K} & \checkmark & \underline{\textbf{1.48}} & \underline{\textbf{3.46}} &	\underline{\textbf{1.81}} & \underline{\textbf{1.63}} \\

\hline
Hirschmuller \cite{hirschmuller2005accurate}  &  - &  - & 8.92 & 20.59  & 10.86 & 9.47 \\ 
Zhou et al. \cite{Zhou_2017_ICCV} & \textcolor{blue}{K} & \checkmark & - & - & 9.91 & - \\
Li and Yuan \cite{Li2018OcclusionAS}  & \textcolor{blue}{K} & \checkmark &  6.89  & 19.42 & 8.98 & 7.39 \\
Tulyakov et al. \cite{tulyakov2017weakly} &  \textcolor{blue}{K} & - &  3.78  & 10.93 & 4.97 & 4.11  \\
Joung et al. \cite{joung2019unsupervised}  & \textcolor{blue}{K} & - & - & -  & 4.47 & - \\
\textbf{Ours}(PSMNet)  & \textcolor{blue}{K} & \checkmark & \textbf{3.13} & \textbf{8.70} & \textbf{4.06} & \textbf{3.86} \\ 
\hline
\end{tabular}
}
\caption{\textbf{KITTI 2015 online benchmark}. We submitted PSMNet, trained on MCN-BM/W-ARC labels, on the KITTI 2015 online stereo benchmark. In \textcolor{blue}{blue} self-supervised methods, while in \textcolor{red}{red} supervised strategies. We indicate with E2E architectures trained in an \ee{} manner, while SF on the SceneFlow dataset \cite{Mayer_2016_CVPR}.}
\label{table:kitti-benchmark}
\end{table}

\begin{figure}[t]
    \centering
    \begin{tabular}{c}
        \includegraphics[width=0.24\textwidth]{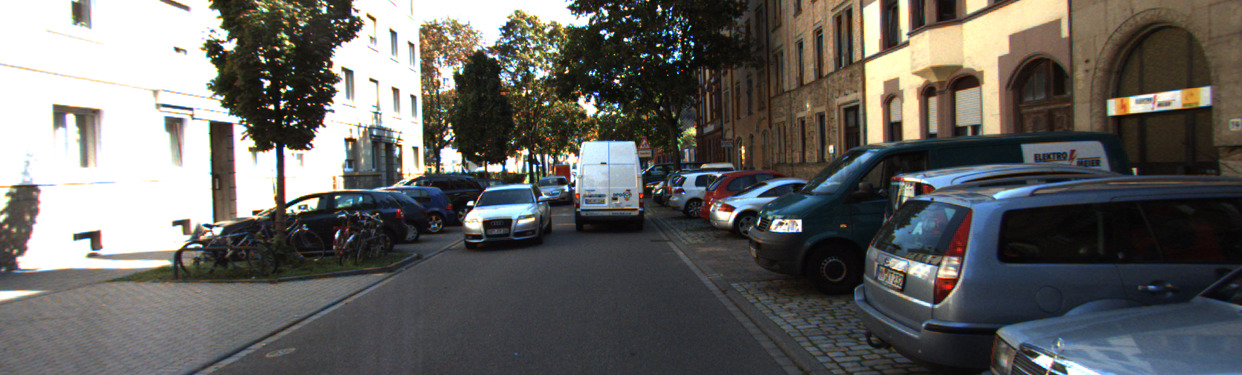} \includegraphics[width=0.24\textwidth]{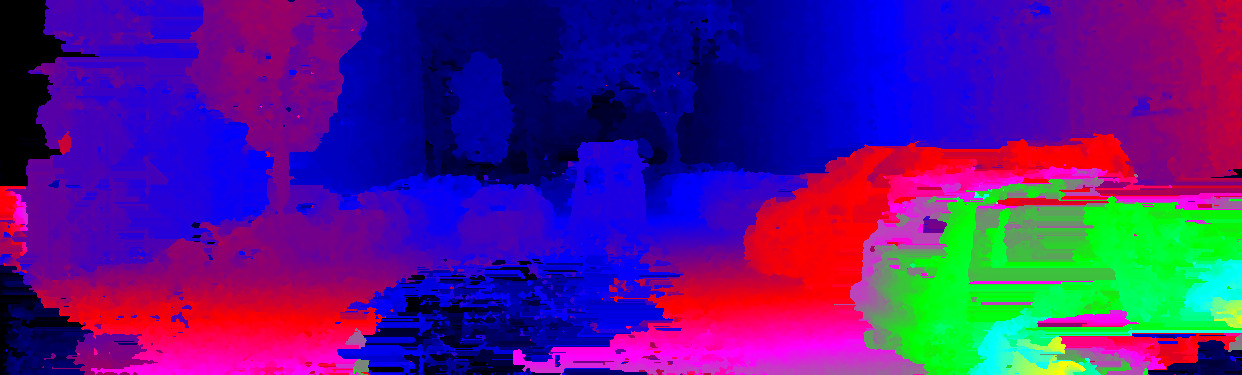} \includegraphics[width=0.24\textwidth]{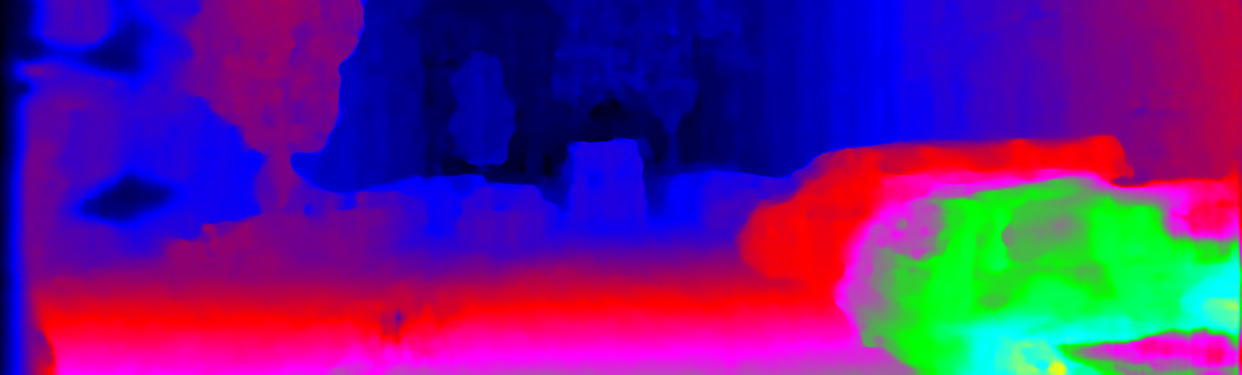} \includegraphics[width=0.24\textwidth]{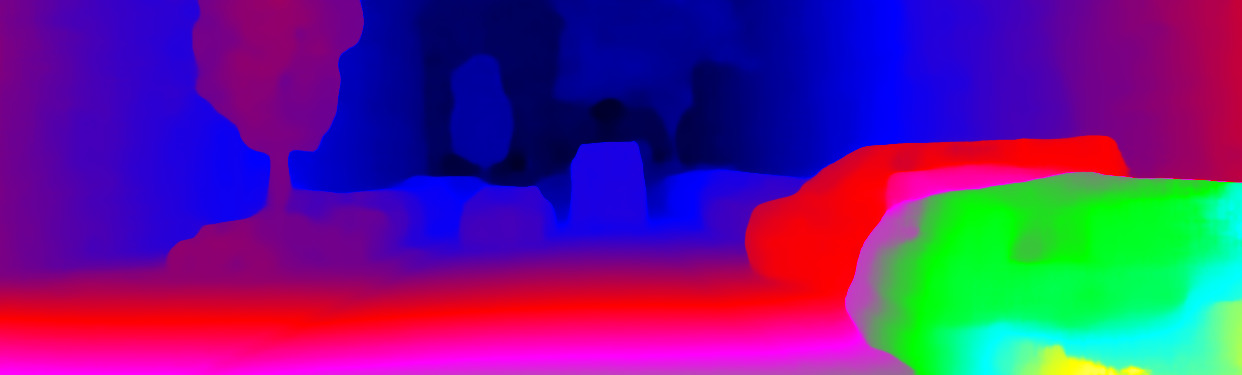} \\
        
        \includegraphics[width=0.24\textwidth]{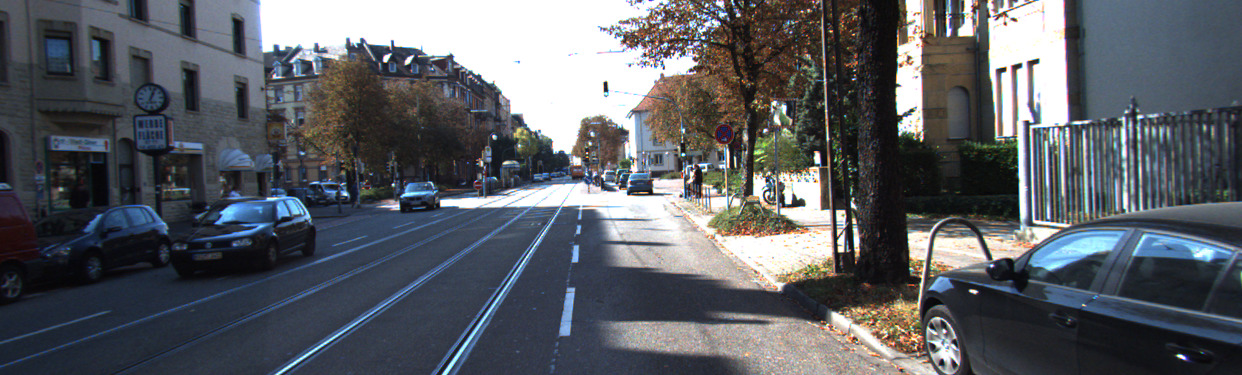} \includegraphics[width=0.24\textwidth]{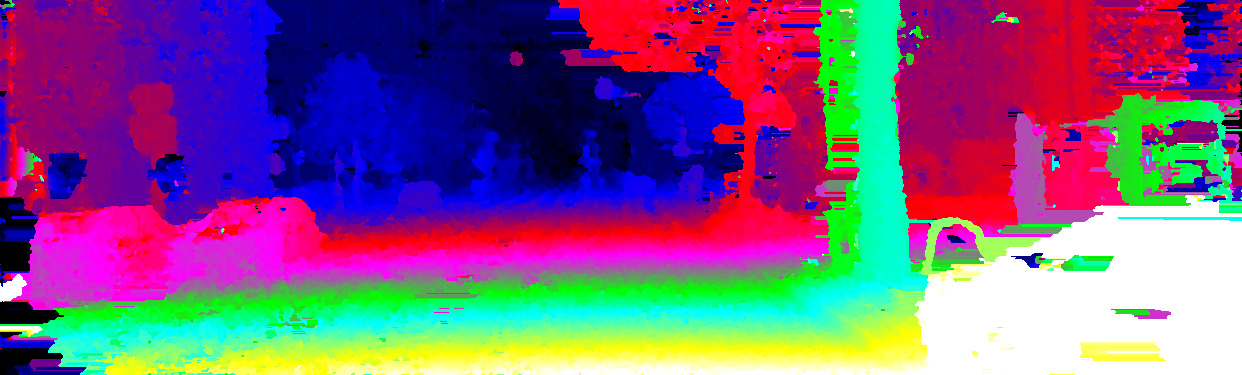} \includegraphics[width=0.24\textwidth]{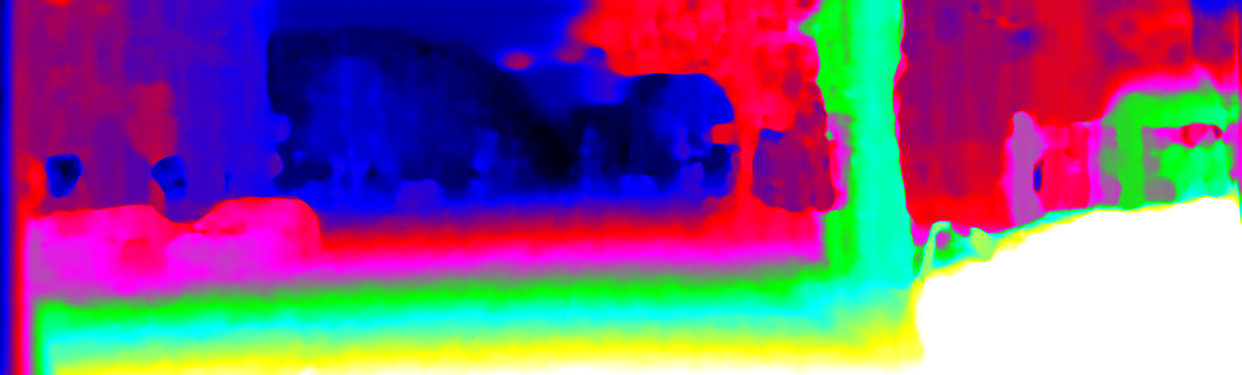} \includegraphics[width=0.24\textwidth]{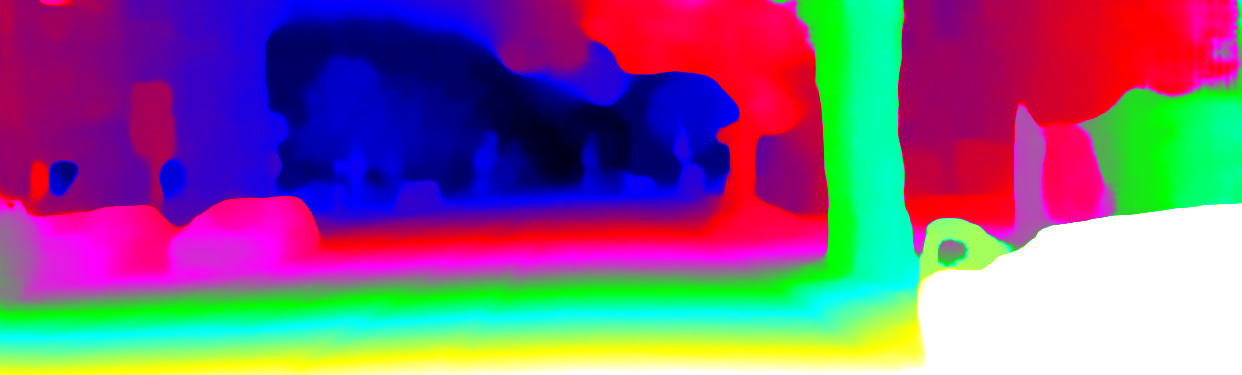} \\
        \includegraphics[width=0.24\textwidth]{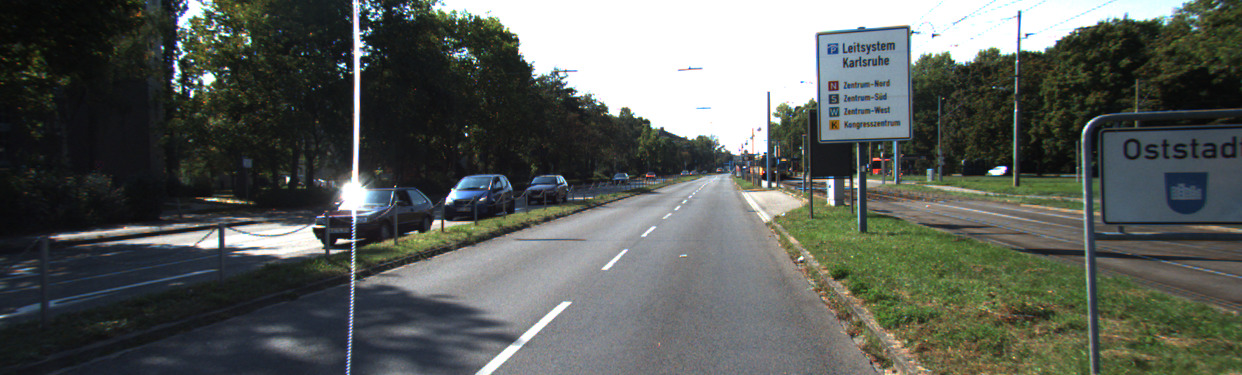}        \includegraphics[width=0.24\textwidth]{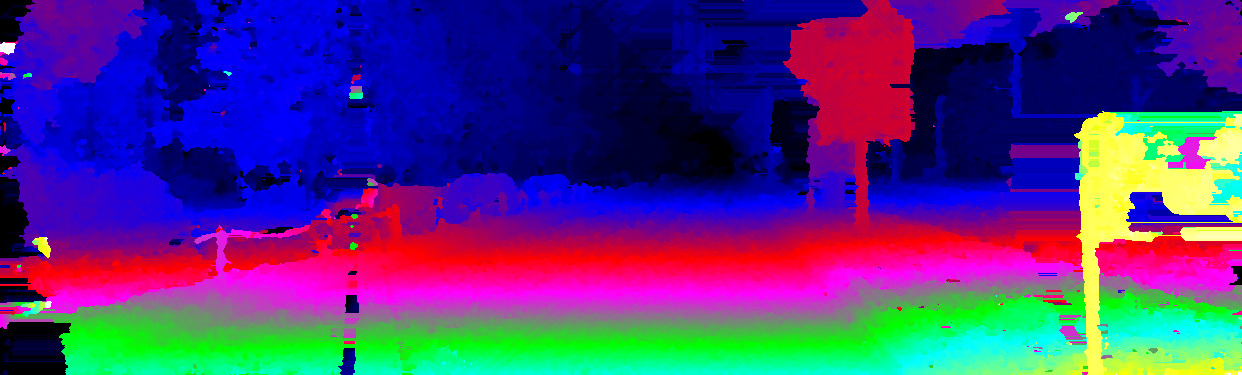} 
        \includegraphics[width=0.24\textwidth]{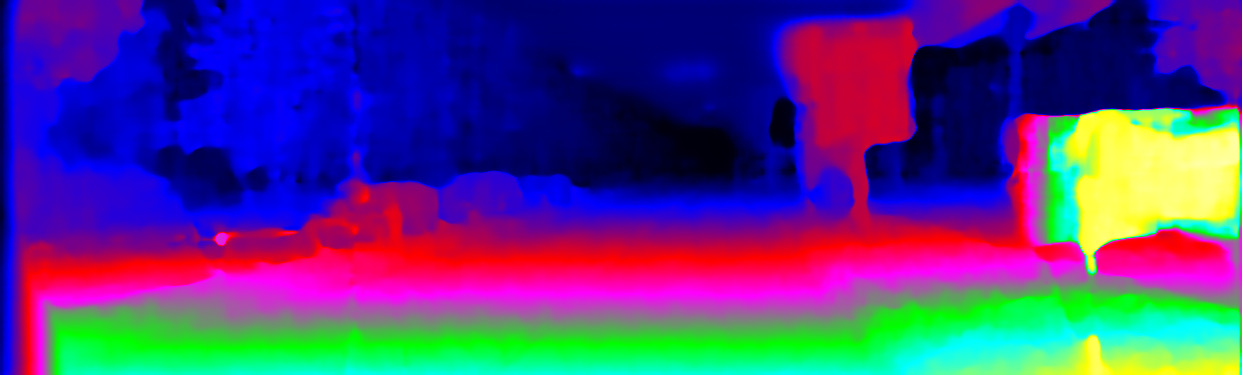} 
        \includegraphics[width=0.24\textwidth]{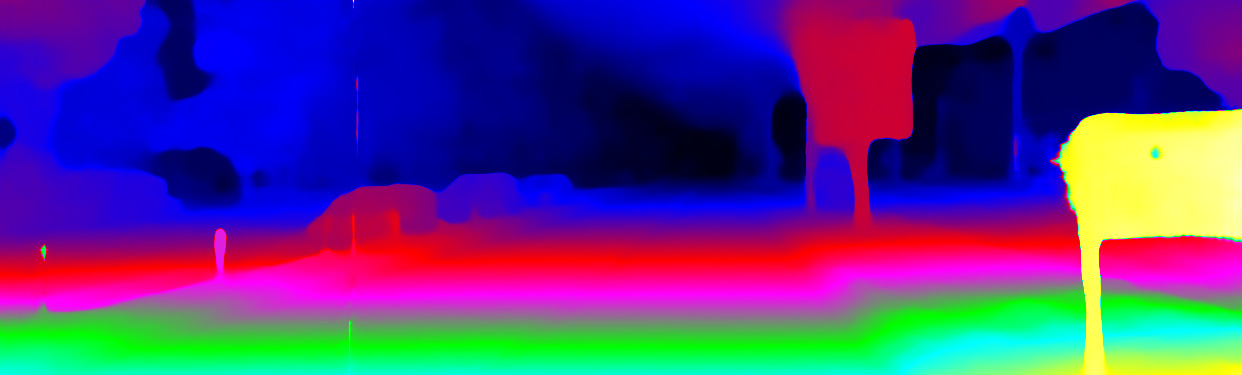} \\
        
    \end{tabular}
    \caption{\textbf{KITTI 2015 online benchmark qualitatives}. From left to right, the reference images, and the disparity maps computed by \cite{SGM_PAMI}, \cite{Li2018OcclusionAS} and our PSMNet trained on MCN-BM/W-ARC labels.}
    \label{fig:kitti_benchmark}
\end{figure}

\subsection{Comparison with state-of-the-art}
We compare our models with state-of-the-art self-supervised stereo methods. Table  \ref{table:comparison_sota} reports, in addition to D1 and EPE, also {RMSE} and {RMSE log} as depth error measurements and $\delta<1.25, \delta<1.25^2, \delta<1.25^3$ accuracy metrics according to \cite{wang2019unos,Zhou_2017_ICCV}. Notice that some of these methods exploit additional information, such as stereo videos \cite{wang2019unos} or adaptation strategies \cite{Zhong_ECCV_2018}. Proxies distilled by MCN-BM/W-ARC can be successfully exploited using both 2D and 3D architectures, enabling even the simplest 2D network \stereodepth{} to outperform all the competitors. Our strategy is effective, allowing all the adopted backbones to improve depth estimation by a notable margin on 6 metrics out of 7. 
Furthermore, we test our PSMNet trained using MCN-BM/W-ARC proxies on the KITTI 2015 online benchmark, reporting the results in Table \ref{table:kitti-benchmark}. Our model not only outperforms \cite{SGM_PAMI} and self-supervised competitors, as can be also perceived in Fig. \ref{fig:kitti_benchmark}, but also supervised strategies \cite{Tonioni_2019_CVPR,Mayer_2016_CVPR} on both non-occluded and all areas.

\subsection{Generalization}

Finally, we show experiments supporting that supervision from our MCN-BM/W-ARC labels achieves good generalization to different domains. To this aim, we run networks trained on KITTI with our paradigm to estimate disparity on Middlebury v3 and ETH3D, framing completely different environments.

%-------------------------------------------------------------------------
% MiddleBury test
%-------------------------------------------------------------------------

\begin{table}[t]
\centering
\scalebox{0.85}{
\setlength{\tabcolsep}{8pt} %% default is 6pt
\begin{tabular}{l|c|cc|cc}
\hline
 Method & Training &  \multicolumn{2}{c|}{Middlebury v3 \cite{scharstein2014high}}
  & \multicolumn{2}{c}{ETH3D \cite{schops2017multi}} \\
  & Dataset & \cellcolor{lower} BAD2 (\%) & \cellcolor{lower} EPE & \cellcolor{lower} BAD2 (\%) &  \cellcolor{lower} EPE\\  
 \hline
Zhang \etal \cite{Zhang2019GANet} & \textcolor{red}{SF+K} & \underline{\textbf{18.90}} & \textbf{3.44} & \underline{\textbf{3.43}} & \textbf{0.91} \\
Chang and Chen \cite{Chang_2018_CVPR} (PSMNet) & \textcolor{red}{SF+K} & 20.04 & 3.01 & 13.07 & 1.35 \\
Guo \etal \cite{guo2019group}(GWCNet) & \textcolor{red}{SF+K} & 21.36 & 3.29 & 19.96 & 1.88 \\

\hline

Wang \etal \cite{wang2019unos}(stereo only) & \textcolor{blue}{K} & 30.55 & 4.77 &  11.17 & 1.47\\
Wang \etal \cite{wang2019unos}(stereo videos) & \textcolor{blue}{K} & 31.63 & 5.23 & 19.59 & 1.97 \\
Lai \etal \cite{lai19cvpr}(stereo videos) & \textcolor{blue}{K} & 45.18 & 6.42 & 10.15 & 1.01  \\

\textbf{Ours}(\stereodepth) & \textcolor{blue}{K} & 27.43 & 3.72 & 6.94 & 1.31\\
\textbf{Ours}(iResNet) & \textcolor{blue}{K} & 25.08 & 3.85 & 6.29 & 0.81 \\
\textbf{Ours}(GWCNet) & \textcolor{blue}{K} &  20.75 & 3.17 & \textbf{3.50} & \underline{\textbf{0.48}} \\
\textbf{Ours}(PSMNet) & \textcolor{blue}{K} & \textbf{19.56} & \underline{\textbf{2.99}} & 4.00  & 0.51 \\
\hline
\end{tabular}

}
\caption{\textbf{Generalization test on Middlebury v3 and ETH3D.} We evaluate networks trained in self-supervised (\textcolor{blue}{blue}) or supervised (\textcolor{red}{red}) fashion on KITTI (K) and SceneFlow dataset (SF) \cite{Mayer_2016_CVPR}.} %\textbf{Ours} indicates stereo networks trained using MCN-BM/W-ARC labels.}
\label{table:middlebury_eth}
\end{table}

\begin{figure}[t]
    \centering
    \renewcommand{\tabcolsep}{0.5pt}
    \begin{tabular}{cccccc}
        \small{Reference} & \small{GT} & \small{\textcolor{blue}{Wang}} \cite{wang2019unos} & \small{\textcolor{blue}{Lai}} \cite{lai19cvpr} & \small{\textcolor{blue}{\textbf{Ours}}} \cite{Chang_2018_CVPR} & \small{\textcolor{red}{Zhang }} \cite{Zhang2019GANet} \\
        \includegraphics[width=0.16\textwidth]{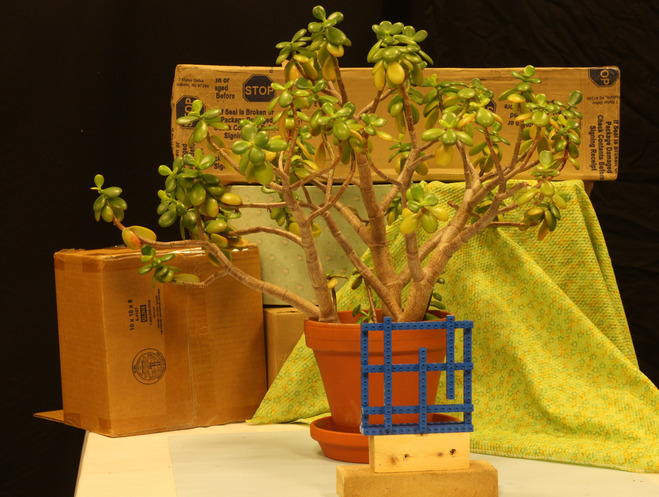} &
        \includegraphics[width=0.16\textwidth]{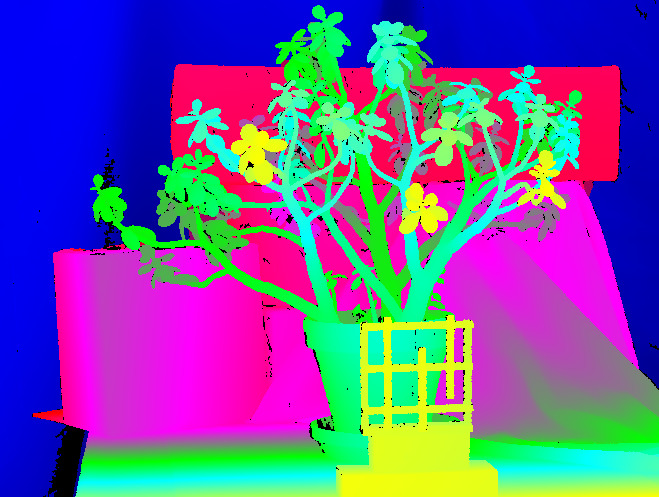} &        
        \includegraphics[width=0.16\textwidth]{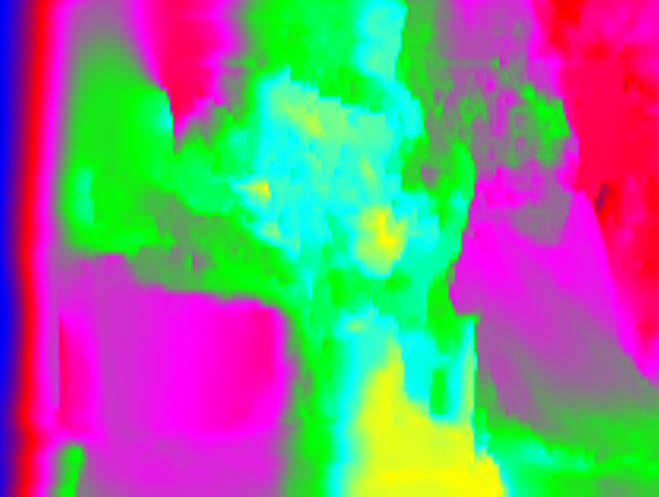} &
        \includegraphics[width=0.16\textwidth]{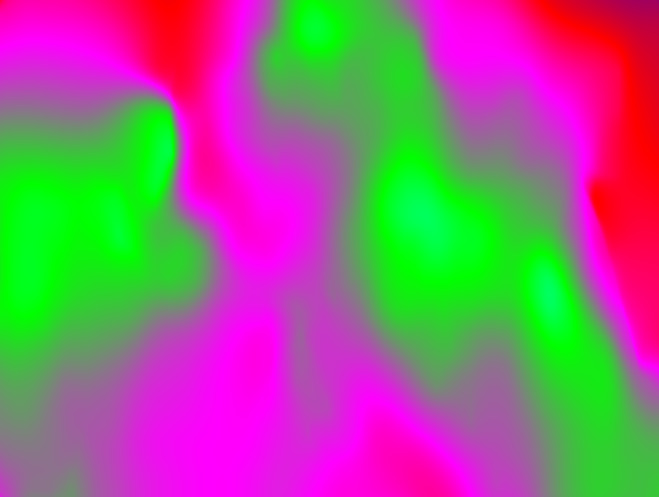} &
        \includegraphics[width=0.16\textwidth]{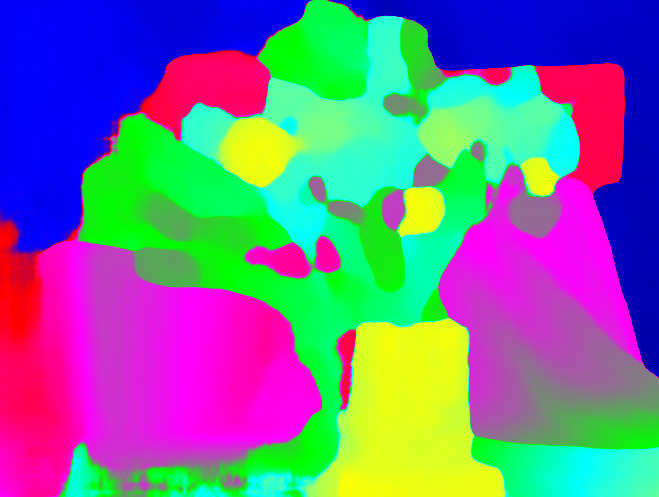} &
        \includegraphics[width=0.16\textwidth]{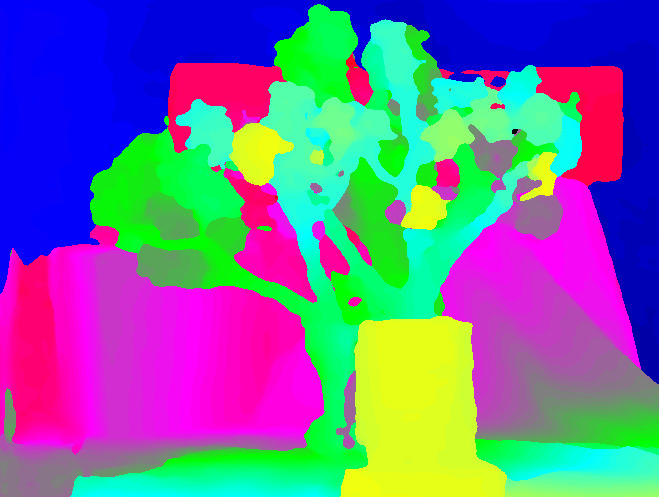}
        \\
        \includegraphics[width=0.16\textwidth]{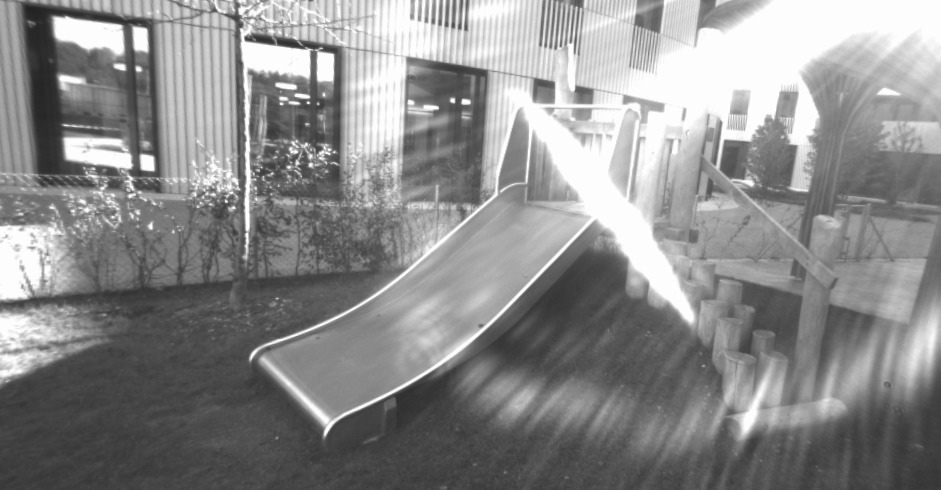} &
        \includegraphics[width=0.16\textwidth]{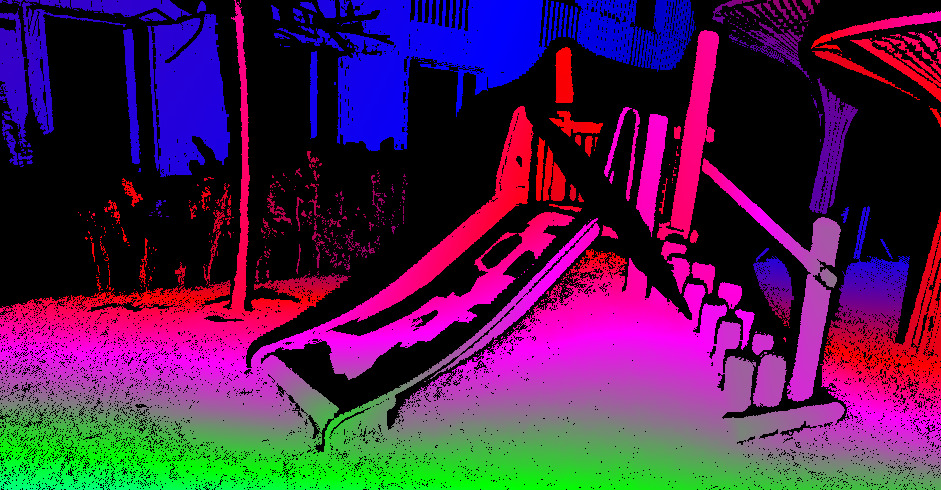} &
        \includegraphics[width=0.16\textwidth]{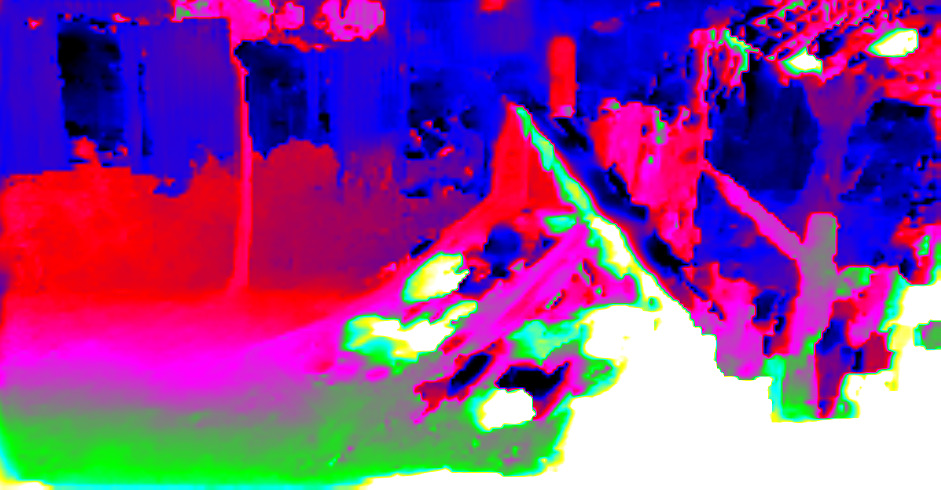} &
        \includegraphics[width=0.16\textwidth]{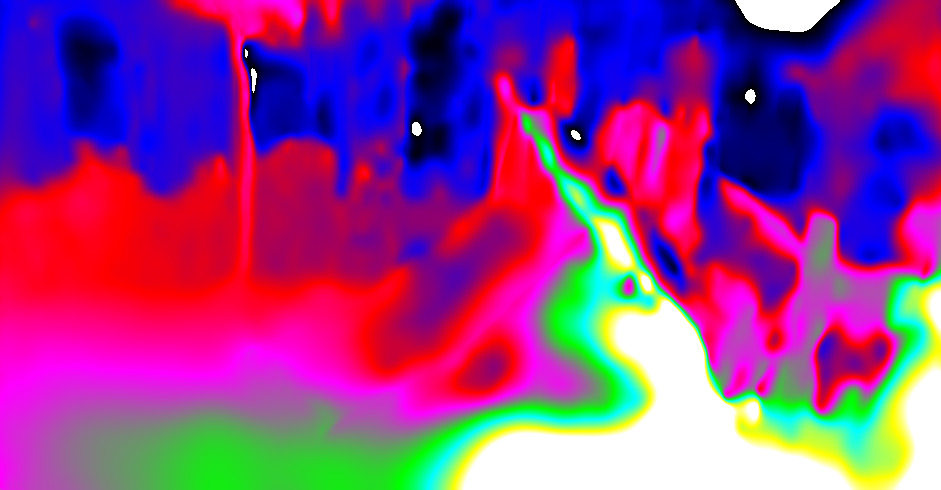} &
        \includegraphics[width=0.16\textwidth]{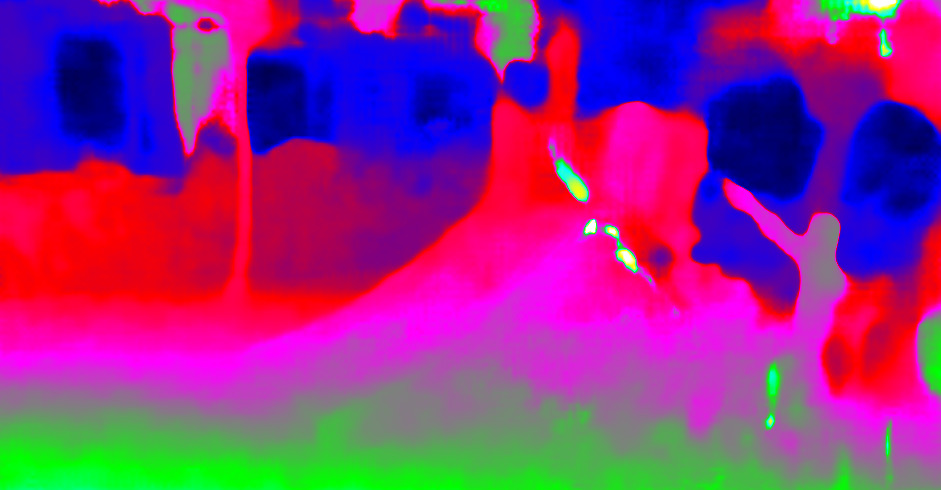} &
        \includegraphics[width=0.16\textwidth]{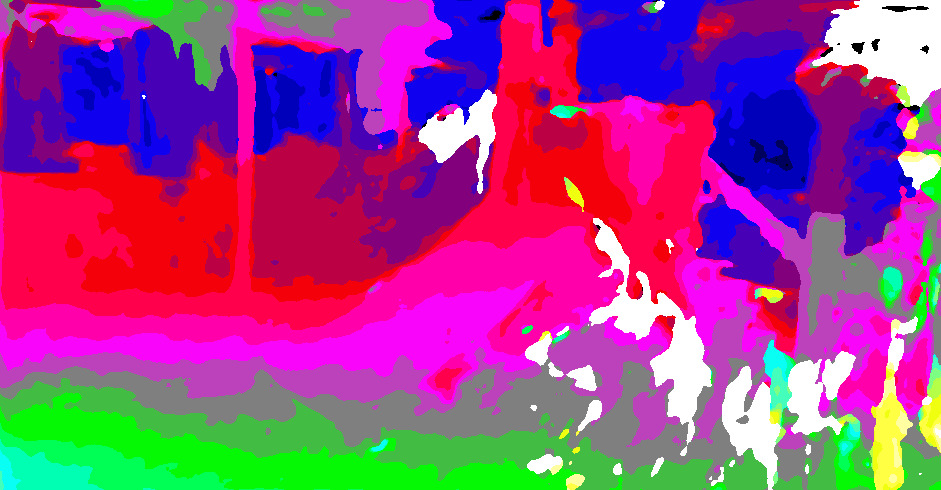}
        \\
    \end{tabular}
    \caption{\textbf{Examples of generalization}. First row shows disparity maps obtained on a stereo pair from the Middlebury v3 dataset, while second from ETH3D. Methods in \textcolor{blue}{\textbf{blue}} are self-supervised, while in \textcolor{red}{\textbf{red}} are supervised with ground-truth. }
    \label{fig:qualitatives_mid}
\end{figure}

Table \ref{table:middlebury_eth} shows the outcome of this evaluation. We report, on top, the performance of fully supervised methods trained on SceneFlow \cite{Mayer_2016_CVPR} and fine-tuned on KITTI for comparison. On bottom, we report self-supervised frameworks trained on the KITTI split from the previous experiments. All networks are transferred without fine-tuning. Compared to existing self-supervised strategies (rows 4-6), networks trained with our proxies achieves much better generalization on both the datasets, performing comparable (or even better) with \gt{} supervised networks.
Fig. \ref{fig:qualitatives_mid} shows few examples from the two datasets, where the structure of the scene is much better recovered when trained on our proxies.

\section{Conclusion}
This paper proposed a novel strategy to source reliable disparity proxy labels in order to train deep stereo networks in a self-supervised manner leveraging a monocular completion paradigm. Well-known stereo artefacts are soften by learning on such labels, that can be obtained from large RGB stereo datasets in which no additional depth information (\eg{} LiDAR or active sensors) is available. Through an extensive ablation study on two popular stereo datasets, we proved that our approach is able to infer accurate yet dense maps starting from points sourced by (potentially) any traditional stereo algorithm, and that such labels provide a strong supervision for both 2D and 3D stereo networks with different complexity. We showed that these networks outperform \sota{} self-supervised methods on KITTI by a large margin and are, in terms of generalization on Middlebury v3 and ETH3D, comparable or even better than \gt{} supervised stereo networks.

\textbf{Acknowledgments.} We gratefully acknowledge the support of NVIDIA Corporation with the donation of the Titan Xp GPU used for this research.

\bibliographystyle{splncs04}
\bibliography{egbib}

\begin{thebibliography}{10}
\providecommand{\url}[1]{\texttt{#1}}
\providecommand{\urlprefix}{URL }
\providecommand{\doi}[1]{https://doi.org/#1}

\bibitem{Chang_2018_CVPR}
Chang, J.R., Chen, Y.S.: Pyramid stereo matching network. In: The IEEE
  Conference on Computer Vision and Pattern Recognition (CVPR). IEEE (June
  2018)

\bibitem{chen2019learning}
Chen, Y., Yang, B., Liang, M., Urtasun, R.: Learning joint 2d-3d
  representations for depth completion. In: IEEE international conference on
  computer vision (ICCV). pp. 10023--10032. IEEE (2019)

\bibitem{Chen_2015_ICCV}
Chen, Z., Sun, X., Wang, L., Yu, Y., Huang, C.: A deep visual correspondence
  embedding model for stereo matching costs. In: The IEEE International
  Conference on Computer Vision (ICCV). IEEE (December 2015)

\bibitem{cheng2018depth}
Cheng, X., Wang, P., Yang, R.: Depth estimation via affinity learned with
  convolutional spatial propagation network. In: European Conference on
  Computer Vision (ECCV). pp. 103--119. Springer (2018)

\bibitem{dovesi2019real}
Dovesi, P.L., Poggi, M., Andraghetti, L., Mart{\'\i}, M., Kjellstr{\"o}m, H.,
  Pieropan, A., Mattoccia, S.: Real-time semantic stereo matching. In: IEEE
  International Conference on Robotics and Automation (ICRA). IEEE (2020)

\bibitem{Eigen_2014}
Eigen, D., Puhrsch, C., Fergus, R.: Depth map prediction from a single image
  using a multi-scale deep network. In: Advances in neural information
  processing systems. pp. 2366--2374. MIT Press (2014)

\bibitem{eldesokey2018propagating}
Eldesokey, A., Felsberg, M., Khan, F.S.: Propagating confidences through cnns
  for sparse data regression. arXiv preprint arXiv:1805.11913  (2018)

\bibitem{Gidaris_2017_CVPR}
Gidaris, S., Komodakis, N.: Detect, replace, refine: Deep structured prediction
  for pixel wise labeling. In: The IEEE Conference on Computer Vision and
  Pattern Recognition (CVPR). IEEE (July 2017)

\bibitem{Godard_CVPR_2017}
Godard, C., {Mac Aodha}, O., Brostow, G.J.: Unsupervised monocular depth
  estimation with left-right consistency. In: IEEE Conference on Computer
  Vision and Pattern Recognition (CVPR). IEEE (2017)

\bibitem{Godard_ICCV_2019}
Godard, C., Mac~Aodha, O., Brostow, G.J.: Digging into self-supervised
  monocular depth estimation. In: IEEE international conference on computer
  vision (ICCV). IEEE (2019)

\bibitem{guo2019group}
Guo, X., Yang, K., Yang, W., Wang, X., Li, H.: Group-wise correlation stereo
  network. In: IEEE Conference on Computer Vision and Pattern Recognition. pp.
  3273--3282. IEEE (2019)

\bibitem{hirschmuller2005accurate}
Hirschmuller, H.: Accurate and efficient stereo processing by semi-global
  matching and mutual information. In: Computer Vision and Pattern Recognition,
  2005. CVPR 2005. IEEE Computer Society Conference on. vol.~2, pp. 807--814.
  IEEE, IEEE (2005)

\bibitem{SGM_PAMI}
Hirschmuller, H.: Stereo processing by semiglobal matching and mutual
  information. IEEE TPAMI  \textbf{30}(2),  328--341 (2008)

\bibitem{huang2019hms}
Huang, Z., Fan, J., Cheng, S., Yi, S., Wang, X., Li, H.: Hms-net: Hierarchical
  multi-scale sparsity-invariant network for sparse depth completion. IEEE
  Transactions on Image Processing  (2019)

\bibitem{ilg2018occlusions}
Ilg, E., Saikia, T., Keuper, M., Brox, T.: Occlusions, motion and depth
  boundaries with a generic network for optical flow, disparity, or scene flow
  estimation. In: 15th European Conference on Computer Vision (ECCV). Springer
  (2018)

\bibitem{joung2019unsupervised}
Joung, S., Kim, S., Park, K., Sohn, K.: Unsupervised stereo matching using
  confidential correspondence consistency. IEEE Transactions on Intelligent
  Transportation Systems  (2019)

\bibitem{Kendall_2017_ICCV}
Kendall, A., Martirosyan, H., Dasgupta, S., Henry, P., Kennedy, R., Bachrach,
  A., Bry, A.: End-to-end learning of geometry and context for deep stereo
  regression. In: The IEEE International Conference on Computer Vision (ICCV).
  IEEE (Oct 2017)

\bibitem{Kingma_2014}
Kingma, D., Ba, J.: Adam: A method for stochastic optimization. arXiv preprint
  arXiv:1412.6980  (2014)

\bibitem{ku2018defense}
Ku, J., Harakeh, A., Waslander, S.L.: In defense of classical image processing:
  Fast depth completion on the cpu. In: 2018 15th Conference on Computer and
  Robot Vision (CRV). pp. 16--22. IEEE, IEEE (2018)

\bibitem{lai19cvpr}
Lai, H.Y., Tsai, Y.H., Chiu, W.C.: Bridging stereo matching and optical flow
  via spatiotemporal correspondence. In: IEEE Conference on Computer Vision and
  Pattern Recognition (CVPR). IEEE (2019)

\bibitem{Laina_3DV_2016}
Laina, I., Rupprecht, C., Belagiannis, V., Tombari, F., Navab, N.: Deeper depth
  prediction with fully convolutional residual networks. In: 3DV. IEEE (2016)

\bibitem{Li2018OcclusionAS}
Li, A., Yuan, Z.: Occlusion aware stereo matching via cooperative unsupervised
  learning. In: ACCV. Springer (2018)

\bibitem{Liang_2018_CVPR}
Liang, Z., Feng, Y., Guo, Y., Liu, H., Chen, W., Qiao, L., Zhou, L., Zhang, J.:
  Learning for disparity estimation through feature constancy. In: The IEEE
  Conference on Computer Vision and Pattern Recognition (CVPR). IEEE (June
  2018)

\bibitem{Liu_IEEE_2016}
Liu, F., Shen, C., Lin, G., Reid, I.: Learning depth from single monocular
  images using deep convolutional neural fields. IEEE Trans. on Pattern
  Analysis and Machine Intelligence  \textbf{38}(10),  2024--2039 (2016)

\bibitem{liu2015depth}
Liu, L.K., Chan, S.H., Nguyen, T.Q.: Depth reconstruction from sparse samples:
  Representation, algorithm, and sampling. IEEE Transactions on Image
  Processing  \textbf{24}(6),  1983--1996 (2015)

\bibitem{luo2016efficient}
Luo, W., Schwing, A.G., Urtasun, R.: Efficient deep learning for stereo
  matching. In: IEEE Conference on Computer Vision and Pattern Recognition. pp.
  5695--5703. IEEE (2016)

\bibitem{ma2019self}
Ma, F., Cavalheiro, G.V., Karaman, S.: Self-supervised sparse-to-dense:
  Self-supervised depth completion from lidar and monocular camera. In: 2019
  International Conference on Robotics and Automation (ICRA). pp. 3288--3295.
  IEEE, IEEE (2019)

\bibitem{Mayer_2016_CVPR}
Mayer, N., Ilg, E., Hausser, P., Fischer, P., Cremers, D., Dosovitskiy, A.,
  Brox, T.: A large dataset to train convolutional networks for disparity,
  optical flow, and scene flow estimation. In: The IEEE Conference on Computer
  Vision and Pattern Recognition (CVPR). IEEE (June 2016)

\bibitem{Menze2015CVPR}
Menze, M., Geiger, A.: Object scene flow for autonomous vehicles. In:
  Conference on Computer Vision and Pattern Recognition (CVPR). IEEE (2015)

\bibitem{Pang_2017_ICCV_Workshops}
Pang, J., Sun, W., Ren, J.S., Yang, C., Yan, Q.: Cascade residual learning: A
  two-stage convolutional neural network for stereo matching. In: The IEEE
  International Conference on Computer Vision (ICCV) Workshops. IEEE (Oct 2017)

\bibitem{PyTorch}
Paszke, A., Gross, S., Massa, F., Lerer, A., Bradbury, J., Chanan, G., Killeen,
  T., Lin, Z., Gimelshein, N., Antiga, L., et~al.: Pytorch: An imperative
  style, high-performance deep learning library. In: Advances in Neural
  Information Processing Systems. pp. 8024--8035. MIT Press (2019)

\bibitem{Poggi_3DV_2018}
Poggi, M., Tosi, F., Mattoccia, S.: Learning monocular depth estimation with
  unsupervised trinocular assumptions. In: 6th International Conference on 3D
  Vision (3DV). IEEE (2018)

\bibitem{scharstein2014high}
Scharstein, D., Hirschm{\"u}ller, H., Kitajima, Y., Krathwohl, G.,
  Ne{\v{s}}i{\'c}, N., Wang, X., Westling, P.: High-resolution stereo datasets
  with subpixel-accurate ground truth. In: German conference on pattern
  recognition. pp. 31--42. Springer, Springer (2014)

\bibitem{scharstein2002taxonomy}
Scharstein, D., Szeliski, R.: A taxonomy and evaluation of dense two-frame
  stereo correspondence algorithms. International journal of computer vision
  \textbf{47}(1-3),  7--42 (2002)

\bibitem{schops2017multi}
Schops, T., Schonberger, J.L., Galliani, S., Sattler, T., Schindler, K.,
  Pollefeys, M., Geiger, A.: A multi-view stereo benchmark with high-resolution
  images and multi-camera videos. In: IEEE Conference on Computer Vision and
  Pattern Recognition. pp. 3260--3269. IEEE (2017)

\bibitem{seki2016}
Seki, A., Pollefeys, M.: Patch based confidence prediction for dense disparity
  map. In: BMVC. BMVA (2016)

\bibitem{Shaked_2017_CVPR}
Shaked, A., Wolf, L.: Improved stereo matching with constant highway networks
  and reflective confidence learning. In: The IEEE Conference on Computer
  Vision and Pattern Recognition (CVPR). IEEE (July 2017)

\bibitem{Smolyanskiy_2018_CVPR_Workshops}
Smolyanskiy, N., Kamenev, A., Birchfield, S.: On the importance of stereo for
  accurate depth estimation: an efficient semi-supervised deep neural network
  approach. In: IEEE Conference on Computer Vision and Pattern Recognition
  (CVPR) Workshops. IEEE (2018)

\bibitem{song2019edgestereo}
Song, X., Zhao, X., Fang, L., Hu, H., Yu, Y.: Edgestereo: An effective
  multi-task learning network for stereo matching and edge detection.
  International Journal of Computer Vision pp. 1--21 (2020)

\bibitem{song2018edgestereo}
Song, X., Zhao, X., Hu, H., Fang, L.: Edgestereo: A context integrated residual
  pyramid network for stereo matching. In: 14th Asian Conference on Computer
  Vision (ACCV). Springer (2018)

\bibitem{Tonioni_2017_ICCV}
Tonioni, A., Poggi, M., Mattoccia, S., Di~Stefano, L.: Unsupervised adaptation
  for deep stereo. In: The IEEE International Conference on Computer Vision
  (ICCV). IEEE (Oct 2017)

\bibitem{tonioni2019unsupervised}
Tonioni, A., Poggi, M., Mattoccia, S., Di~Stefano, L.: Unsupervised domain
  adaptation for depth prediction from images. IEEE Transactions on Pattern
  Analysis and Machine Intelligence  (2019)

\bibitem{Tonioni_2019_learn2adapt}
Tonioni, A., Rahnama, O., Joy, T., Di~Stefano, L., Thalaiyasingam, A., Torr,
  P.: Learning to adapt for stereo. In: The IEEE Conference on Computer Vision
  and Pattern Recognition (CVPR). IEEE (June 2019)

\bibitem{Tonioni_2019_CVPR}
Tonioni, A., Tosi, F., Poggi, M., Mattoccia, S., Stefano, L.D.: Real-time
  self-adaptive deep stereo. In: The IEEE Conference on Computer Vision and
  Pattern Recognition (CVPR). IEEE (June 2019)

\bibitem{Tosi_CVPR_2019}
Tosi, F., Aleotti, F., Poggi, M., Mattoccia, S.: Learning monocular depth
  estimation infusing traditional stereo knowledge. In: The IEEE Conference on
  Computer Vision and Pattern Recognition (CVPR). IEEE (2019)

\bibitem{BMVC_2017}
Tosi, F., Poggi, M., Tonioni, A., Di~Stefano, L., Mattoccia, S.: Learning
  confidence measures in the wild. In: BMVC. BMVA (Sept 2017)

\bibitem{tulyakov2017weakly}
Tulyakov, S., Ivanov, A., Fleuret, F.: Weakly supervised learning of deep
  metrics for stereo reconstruction. In: IEEE Conference on Computer Vision and
  Pattern Recognition. pp. 1339--1348. IEEE (2017)

\bibitem{Uhrig2017THREEDV}
Uhrig, J., Schneider, N., Schneider, L., Franke, U., Brox, T., Geiger, A.:
  Sparsity invariant cnns. In: International Conference on 3D Vision (3DV).
  IEEE (2017)

\bibitem{wang2019unos}
Wang, Y., Wang, P., Yang, Z., Luo, C., Yang, Y., Xu, W.: Unos: Unified
  unsupervised optical-flow and stereo-depth estimation by watching videos. In:
  IEEE Conference on Computer Vision and Pattern Recognition. pp. 8071--8081.
  IEEE (2019)

\bibitem{Watson_ICCV_2019}
Watson, J., Firman, M., Brostow, G.J., Turmukhambetov, D.: Self-supervised
  monocular depth hints. In: IEEE international conference on computer vision
  (ICCV). IEEE (2019)

\bibitem{Watson_ECCV_2020}
Watson, J., Mac~Aodha, O., Turmukhambetov, D., Brostow, G.J., Firman, M.:
  Learning stereo from single images. In: European Conference on Computer
  Vision (ECCV). Springer (2020)

\bibitem{yang2019drivingstereo}
Yang, G., Song, X., Huang, C., Deng, Z., Shi, J., Zhou, B.: Drivingstereo: A
  large-scale dataset for stereo matching in autonomous driving scenarios. In:
  IEEE Conference on Computer Vision and Pattern Recognition (CVPR). IEEE
  (2019)

\bibitem{yang2018segstereo}
Yang, G., Zhao, H., Shi, J., Deng, Z., Jia, J.: Segstereo: Exploiting semantic
  information for disparity estimation. In: 15th European Conference on
  Computer Vision (ECCV). Springer (2018)

\bibitem{yang2007spatial}
Yang, Q., Yang, R., Davis, J., Nist{\'e}r, D.: Spatial-depth super resolution
  for range images. In: 2007 IEEE Conference on Computer Vision and Pattern
  Recognition. pp.~1--8. IEEE, IEEE (2007)

\bibitem{yu2018deep}
Yu, L., Wang, Y., Wu, Y., Jia, Y.: Deep stereo matching with explicit cost
  aggregation sub-architecture. In: Thirty-Second AAAI Conference on Artificial
  Intelligence. AAAI Press (2018)

\bibitem{Secaucus_1994_ECCV}
Zabih, R., Woodfill, J.: Non-parametric local transforms for computing visual
  correspondence. In: Third European Conference on Computer Vision (Vol. II).
  pp. 151--158. 3rd European Conference on Computer Vision (ECCV),
  Springer-Verlag New York, Inc., Secaucus, NJ, USA (1994)

\bibitem{zbontar2016stereo}
Zbontar, J., LeCun, Y.: Stereo matching by training a convolutional neural
  network to compare image patches. Journal of Machine Learning Research
  \textbf{17}(1-32), ~2 (2016)

\bibitem{Zhang2019GANet}
Zhang, F., Prisacariu, V., Yang, R., Torr, P.H.: Ga-net: Guided aggregation net
  for end-to-end stereo matching. In: IEEE Conference on Computer Vision and
  Pattern Recognition. pp. 185--194. IEEE (2019)

\bibitem{ZhongArxiv2017}
Zhong, Y., Li, H., Dai, Y.: Self-supervised learning for stereo matching with
  self-improving ability. arXiv preprint arXiv:1709.00930  (2017)

\bibitem{Zhong_ECCV_2018}
Zhong, Y., Li, H., Dai, Y.: Open-world stereo video matching with deep rnn. In:
  ECCV. Springer (2018)

\bibitem{Zhou_2017_ICCV}
Zhou, C., Zhang, H., Shen, X., Jia, J.: Unsupervised learning of stereo
  matching. In: The IEEE International Conference on Computer Vision (ICCV).
  IEEE (October 2017)

\bibitem{Zhou_2017_CVPR}
Zhou, T., Brown, M., Snavely, N., Lowe, D.G.: Unsupervised learning of depth
  and ego-motion from video. In: The IEEE Conference on Computer Vision and
  Pattern Recognition (CVPR). IEEE (July 2017)

\end{thebibliography}

%\newpage\phantom{Supplementary}
%\multido{\i=1+1}{11}{
%	\includepdf[page={\i}]{supplementary.pdf}
%}

\end{document}